\documentclass[10pt,journal,cspaper,compsoc]{IEEEtran}
\usepackage[cmex10]{amsmath}
\usepackage{subfig}

\usepackage{url}

\usepackage{epsfig}
\usepackage{graphicx}
\usepackage{amsmath}
\usepackage{amssymb}
\usepackage{color}

\graphicspath{{./}{./figs/}}

\begin{document}

\title{A Generalized Probabilistic Framework for Compact Codebook Creation}

\author{Lingqiao Liu,
        Lei Wang,
        Chunhua Shen 
\IEEEcompsocitemizethanks{\IEEEcompsocthanksitem Lingqiao Liu is with the college of engineering and computer science, the Australian National University, Canberra, Australia, ACT 0602.\protect\\
E-mail: liulq83@gmail.com
\IEEEcompsocthanksitem Lei Wang is with the School of Computer Science and
Software Engineering, University of Wollongong, Room 219, Building 3,
Northfields Avenue, Wollongong, NSW 2500, Australia.\protect\\
E-mail:leiw@uow.edu.au
\IEEEcompsocthanksitem Chunhua Shen is with the School of Computer Science, The University of
Adelaide, Adelaide, SA 5005, Australia. \protect\\ E-mail: chhshen@gmail.com
}
\thanks{}}

\IEEEcompsoctitleabstractindextext{%
\begin{abstract}
  Compact and discriminative visual codebooks are preferred in many visual recognition tasks. In the literature, a number of works have taken the approach of hierarchically merging visual words of an initial large-sized codebook, but implemented this approach with different merging criteria. In this work, we propose a single probabilistic framework to unify these merging criteria, by identifying two key factors: the function used to model class-conditional distribution and the method used to estimate the distribution parameters. More importantly, by adopting new distribution functions and/or parameter estimation methods, our framework can readily produce a spectrum of novel merging criteria. Three of them are specifically focused in this work. In the first criterion, we adopt the multinomial distribution with Bayesian method; In the second criterion, we integrate Gaussian distribution with maximum likelihood parameter estimation. In the third criterion, which shows the best merging performance, we propose a max-margin-based parameter estimation method and apply it with multinomial distribution. Extensive experimental study is conducted to systematically analyse the performance of the above three criteria and compare them with existing ones. As demonstrated, the best criterion obtained in our framework achieves the overall best merging performance among the comparable merging criteria developed in the literature.

\end{abstract}

}

\maketitle

\IEEEdisplaynotcompsoctitleabstractindextext

\section{Introduction}
    In the past few years, the bag-of-words (BoW) model has gained its popularity in visual recognition thanks to its simplicity and efficiency \cite{Csurka-ECCVWS-04,CreatingCodebook,Laptev08,VideoGoogle}. It usually works as follows: A set of local patches (for still images) or local spatial-temporal volumes (for videos) are extracted and represented by local descriptors. These descriptors are processed, for example, by $k$-means clustering~\cite{Csurka-ECCVWS-04}, to form a collection of visual words, which in turn forms a visual codebook. By assigning each local descriptor to the closest (or multiple) visual word(s), a histogram indicating the number of occurrences of each visual word is obtained to characterise an image or video sequence. Among all the factors of the BoW model, visual codebook plays a pivotal role in determining recognition performance. Usually, a sufficiently large-sized codebook (for example, up to thousands of visual words) has to be used to ensure satisfactory recognition performance.

    However, a large-sized codebook can be unfavourable in some cases. For example, as indicated in \cite{smartdictionary}, when localising an object in an image, the computational cost and memory requirement for generating the histogram of each candidate window is proportional to codebook size. To model the interaction between visual words, the pair-wise relationship among visual words is considered in \cite{MMI}. However, the number of pairs quadratically increases with codebook size. In addition, a large-sized codebook leads to high-dimensional image representation, which could make many machine learning algorithms become inefficient and unreliable or even breakdown. Nevertheless, simply reducing the value of $k$ in $k$-means clustering will quickly degrade recognition performance due to the loss of discriminative information. To handle this situation, one of the effective approaches in the literature is to hierarchically merge visual words of an initial large-sized codebook while minimising the loss of discriminative information in the whole course~\cite{MoosmannNIPS06,LazebnikTPAMI08,WangCVPR10}. In this paper, we focus on this method and call it ``word-merging'' in short in the following parts.  

    Essentially, word-merging can be regarded as a dimensionality reduction method. However, comparing with general-purpose dimensionality reduction methods, word-merging methods enjoy two major advantages: i) the speed of performing dimensionality reduction by merging words is much faster. Let $D$ and $d$ be the dimension of the original image representation and the targeted dimension, respectively. The computational cost of word-merging is merely $\mathcal{O}(D)$, which corresponds to a linear scan of the $D$ dimensions. This is in sharp contrast to $\mathcal{O}(Dd)$ as required in commonly used linear-projection-based dimensionality reduction methods. This advantage makes word-merging an attractive  option in computation- or memory-sensitive applications, such as object detection in \cite{smartdictionary}; ii) unlike linear-projection-methods which merge all dimensions via a weighted linear combination,  word-merging methods partition all dimensions into mutually exclusive clusters and then combine them. This process well maintains the ``visual word'' concept, which is important when modelling spatial relationship between visual words \cite{MMI} or visualizing ``discriminative visual words'' is needed.

    In the literature, a number of previous studies have implemented the idea of hierarchical visual word merging with different models and criteria. In~\cite{MMI,smartdictionary}, the mutual information between words and class labels is used to identify the optimal pair of words to merge at each level of the hierarchy. In~\cite{Fast2008Wang}, the scatter-matrix-based class separability is taken as a criterion to seek the optimal pair of words to merge. The work of \cite{UVD} differs from the previous work in that a more rigorous probabilistic model is used to merge visual words. In their work, the optimal pair is sought as the one after which is merged, the resulting histograms can maximize the posterior probability of true class labels. Nevertheless, as reported in~\cite{smartdictionary,Fast2008Wang}, the merging criterion of \cite{UVD} often produces results inferior to those in~\cite{smartdictionary,Fast2008Wang}. This is in a sharp contrast to the expected power of a rigorous probabilistic model.

    In this work, we follow the basic probabilistic model in~\cite{UVD} and discuss its two key factors: the function used to model class-conditional distribution and the method used to estimate the distribution parameters. The difference between our work and \cite{UVD} is that the two key factors are fixed in \cite{UVD} whereas they are treated as flexible components in our work. As will be seen, such a difference is critical because varying these two factors could bring forth markedly different characteristics to the probabilistic model.
    By properly choosing different settings to the two factors, \textit{we achieve a generalized probabilistic framework for merging visual words}.
    With our framework, we show that existing merging criteria can be viewed as the special cases of the probabilistic model, with different combinations of class-conditional distributions and parameter estimation methods. More importantly, through exploring new combinations of class-conditional distribution and parameter estimation method, we are able to produce a spectrum of new merging criteria. In particular, three of them are explored in this work. The first one adopts the same parameter estimation method (Bayesian method) in \cite{UVD} but replaces its distribution model with multinomial distribution. The second one combines a Gaussian distribution with maximum likelihood parameter estimation. In the third merging criterion, we propose a max-margin-based parameter estimation method and apply it with multinomial distribution. Through extensive experimental study, we compare the performance of various merging criteria and analyse their differences. Moreover, we show that the third merging criterion produced by our framework achieves the overall best performance among the comparable algorithms in the literature. 

In sum, this work has made the following contributions:
    \begin{itemize}
    \item{By employing appropriate distribution functions and parameter estimation methods, our generalized probabilistic framework reproduces the criteria in~\cite{smartdictionary} and~\cite{Fast2008Wang} as special cases;}
    \item{With this framework, we propose a new criterion by modelling each class with a multinomial distribution function. It can achieve better recognition performance than that originally proposed in \cite{UVD}.}
    \item{With this framework, we explore the combination of Gaussian distribution and maximum likelihood estimation to produce another merging criterion;}
    \item{Based on this framework, we put forward a max-margin-based parameter estimation method, leading to another new criterion. It gives the overall highest recognition performance when compared with all the above word-merging criteria.}
    \end{itemize}

\section{Related Work}\label{sec:related-work}
    This section reviews the supervised compact codebook creation methods  in \cite{smartdictionary,Fast2008Wang,UVD}, with the focus on \cite{UVD} which inspires our work. As shown in~\cite{Fast2008Wang}, compact codebook creation can essentially be casted as a large-scale discrete optimization problem, subject to a criterion related to the discriminative power of the resultant compact codebook. Due to the difficulty of efficient and global optimization, hierarchically merging visual words is often adopted in the literature. That is, two words are identified at each level of the hierarchy such that merging them will optimize a given criterion. Let ${\mathcal B}^{t+1}$ denote a visual codebook consisting of $t+1$ words. Let ${\mathcal B}^{t}_{r,s}$ be the resultant codebook after merging the $r$th and $s$th words. The corresponding histogram for the $i$th sample is denoted by ${\mathbf h}^{t}_i$, and its $j$th bin is $h^{t}_{ij}$, where $1\leq{i}\leq{n},~~1\leq{j}\leq{t}$. Also, $c\in\{1,2,\cdots,C\}$ is the class label of a training sample. In this paper, the criteria in \cite{smartdictionary,Fast2008Wang,UVD} are termed AIB, CSM and UVD in short, respectively.

    \textbf{AIB}: In \cite{smartdictionary}, the mutual information, $I$, between ${\mathcal B}^{t}_{r,s}$ and class labels $c$ is used to measure its discriminative power as
    \begin{equation}\label{eqn:FulkersonECCV08-MMI}
        I({\mathcal B}^{t}_{r,s},c)=\sum_{j=1}^{t}\sum_{c=1}^{C}P(v^{t}_j,c)\log\frac{P(v^{t}_j,c)}{P(v^{t}_j)P(c)},
    \end{equation}where $v^{t}_j$ denotes the $j$th word of ${\mathcal B}_{r,s}^{t}$ and $P(v^{t}_j,c)$ and $P(v^{t}_j)$ are estimated with the $j$th bins of training histograms. At each level $t$, the words $r$ and $s$ whose mergence maximizes $I({\mathcal B}^{t}_{r,s},c)$ are identified and merged. As noted in \cite{smartdictionary}, this criterion can be related to agglomerative information bottleneck~\cite{SlonimNIPS99}.

    \textbf{CSM}: In \cite{Fast2008Wang}, the scatter-matrix-based class separability, $S$, is used to measure the goodness of ${\mathcal B}^{t}_{r,s}$ as
    \begin{equation}
    S(r,s) = {\mathrm{tr}({\mathbf S}_{w})}/{\mathrm{tr}({\mathbf S}_{t})},
    \end{equation}where ${\mathbf S}_{w}$ and ${\mathbf S}_{t}$ are the within-class scatter matrix and the total scatter matrix, respectively. ${\mathrm {tr}}(\cdot)$ denotes the trace of a matrix. They are computed with training histograms ${\mathbf h}^{t}_1,\cdots,{\mathbf h}^{t}_n$. At each level, the words $r$ and $s$ whose mergence minimizes $S(r,s)$ are identified and merged~\footnote{To facilitate the subsequent analysis, we use the minimization of ${\mathrm{tr}({\mathbf S}_{w})}/{\mathrm{tr}({\mathbf S}_{t})}$ here. Because of the identity ${\mathrm{tr}({\mathbf S}_{t})}={\mathrm{tr}({\mathbf S}_{b})}+{\mathrm{tr}({\mathbf S}_{w})}$, it is equivalent to \cite{Fast2008Wang} which maximizes ${\mathrm{tr}({\mathbf S}_{b})}/{\mathrm{tr}({\mathbf S}_{t})}$.}.

    \textbf{UVD}: In~\cite{UVD}, the posterior probability of true class labels conditioned on ${\mathcal B}^{t}_{r,s}$ is proposed to measure the discriminative power of ${\mathcal B}^{t}_{r,s}$. Let $\hat{\mathbf c}=\{{c}_1,\cdots,{c}_n\}$ be the label set of the $n$ training samples. Let ${\mathcal H}^{t}=\{{\mathbf h}^{t}_1,\cdots,{\mathbf h}^{t}_n\}$ be the set of $n$ training histograms obtained with ${\mathcal B}^{t}_{r,s}$. Using the Bayes' theorem, this posterior probability is computed as
    \begin{equation}\label{FrameworkT}
    P(\hat{\mathbf c}|{\mathcal H}^{t}) = \frac{P({\mathcal H}^{t}|\hat{\mathbf c})P(\hat{\mathbf c})}{\sum_{{\mathbf c}'}P({\mathcal H}^{t}|{\mathbf c}') P({\mathbf c}')},
    \end{equation}where $P({\mathcal H}^{t}|\hat{\mathbf c})$ is the likelihood of the $n$ training histograms conditioned on true label configuration $\hat{\mathbf c}$, and $P({\mathcal H}^{t}|{\mathbf c}')$ is the likelihood conditioned on any one of $C^{n}$ possible label configurations. Due to the difficulty of enumerating all possible configurations, \cite{UVD} approximates the denominator with two configurations only: the true configuration $\hat{\mathbf c}$ and a special configuration ${\mathbf c}^{\mathrm{same}}$ in which all training samples have a same class label. Assuming equal prior over these two configurations, it gives:
     \begin{eqnarray}\label{FrameworkP}
        P(\hat{\mathbf c}|{\mathcal H}^{t})
        & \approx & \frac{P({\mathcal H}^{t}|\hat{\mathbf c})}{P({\mathcal H}^{t}|\hat{\mathbf c}) + P({\mathcal H}^{t}|{\mathbf c}^{\mathrm{same}}) }. \nonumber \\
    \end{eqnarray}Thus, maximizing $P(\hat{\mathbf c}|{\mathcal H}^{t})$ is (approximately) equivalent to maximizing $\frac{P({\mathcal H}^{t}|\hat{\mathbf c})}{P({\mathcal H}^{t}|{\mathbf c}^{\mathrm{same}})}$. The likelihood $P({\mathcal H}^{t}|{\mathbf c})$ is computed as
    \begin{equation}\label{eqn:BayesianModel}
    P({\mathcal H}^{t}|{\mathbf c})=\prod_{c=1}^{C}\int\prod_{{\mathbf h}^{t}_i\in{{\mathcal D}_c}}P({\mathbf h}^{t}_{i}|\boldsymbol\theta_{c})P(\boldsymbol\theta_{c})d\boldsymbol\theta_{c}
    \end{equation}where $P({\mathbf h}^{t}_{i}|\boldsymbol\theta_{c})$
    is the class-conditional distribution for class $c$, $\boldsymbol\theta_{c}$ its parameter set, and ${\mathcal D}_c$ the set of all training samples in class $c$. In~\cite{UVD}, $P({\mathbf h}^{t}_{i}|\boldsymbol\theta_{c})$ is modeled as a Gaussian distribution\footnote{As suggested in~\cite{UVD}, the square root of each bin of ${\mathbf h}$ is used to better fit the Gaussian distribution assumption.}.
    A conjugate Gaussian-gamma prior is defined over $\boldsymbol{\theta}_c$ as $P(\boldsymbol{\theta}_c|\mu,\lambda,a,b)$, where $\mu$, $\lambda$, $a$, and $b$ are hyper-parameters. Assuming the independence of different bins and i.i.d samples in each class, the above likelihood is obtained as
    \begin{equation}\label{}
    P({\mathcal H}^{t}|{\mathbf c})=\prod_{c=1}^{C}\prod_{j=1}^{t}\int\prod_{{\mathbf h}^{t}_i\in{{\mathcal D}_c}}P({h}^{t}_{ij}|\theta_{cj})P(\theta_{cj})d\theta_{cj},
    \end{equation}where $h^{t}_{ij}$ is the $j$th bin of the histogram ${\mathbf h}^{t}_{i}$, and $\theta_{ci}$ is the parameter set (mean and variance) for the $j$th bin in class $c$. Since $P(\theta_{cj})$ is the conjugate prior of $P(h^{t}_{ij}|\theta_{cj})$, the integral can be analytically worked out. At each level of the hierarchy, the words $r$ and $s$ whose mergence maximizes ${P({\mathcal H}^{t}|\hat{\mathbf c})}/{P({\mathcal H}^{t}|{\mathbf c}^{\mathrm{same}})}$ is identified and merged.


\section{The proposed Generalized Probabilistic Framework}
    In this paper, we take the basic formulation in Eq.(\ref{FrameworkP}) and develop it to a general probabilistic framework. Any algorithm taking such a formulation needs to determine \textit{two key factors: i) how to model the class-conditional distribution $P({\mathbf h}_{i}|\boldsymbol\theta_{c})$ in Eq.(\ref{eqn:BayesianModel})~\footnote{In this section, we drop the superscript $t$ in ${\mathbf h}^{t}_{i}$. All the calculation is now at the level $t$ unless indicated otherwise.}; ii) how to estimate the model parameter $\boldsymbol\theta_{c}$}. As shown in Section~\ref{sec:related-work}, UVD \cite{UVD} models $P({\mathbf h}_{i}|\boldsymbol\theta_{c})$ with a Gaussian distribution and uses the Bayesian method to marginalize out the model parameter $\boldsymbol\theta_{c}$. The effect of $\boldsymbol\theta_{c}$ is averaged with a Gaussian-gamma prior and its value is not explicitly estimated.
    \begin{figure}
    \centering
            \includegraphics[height=40mm]{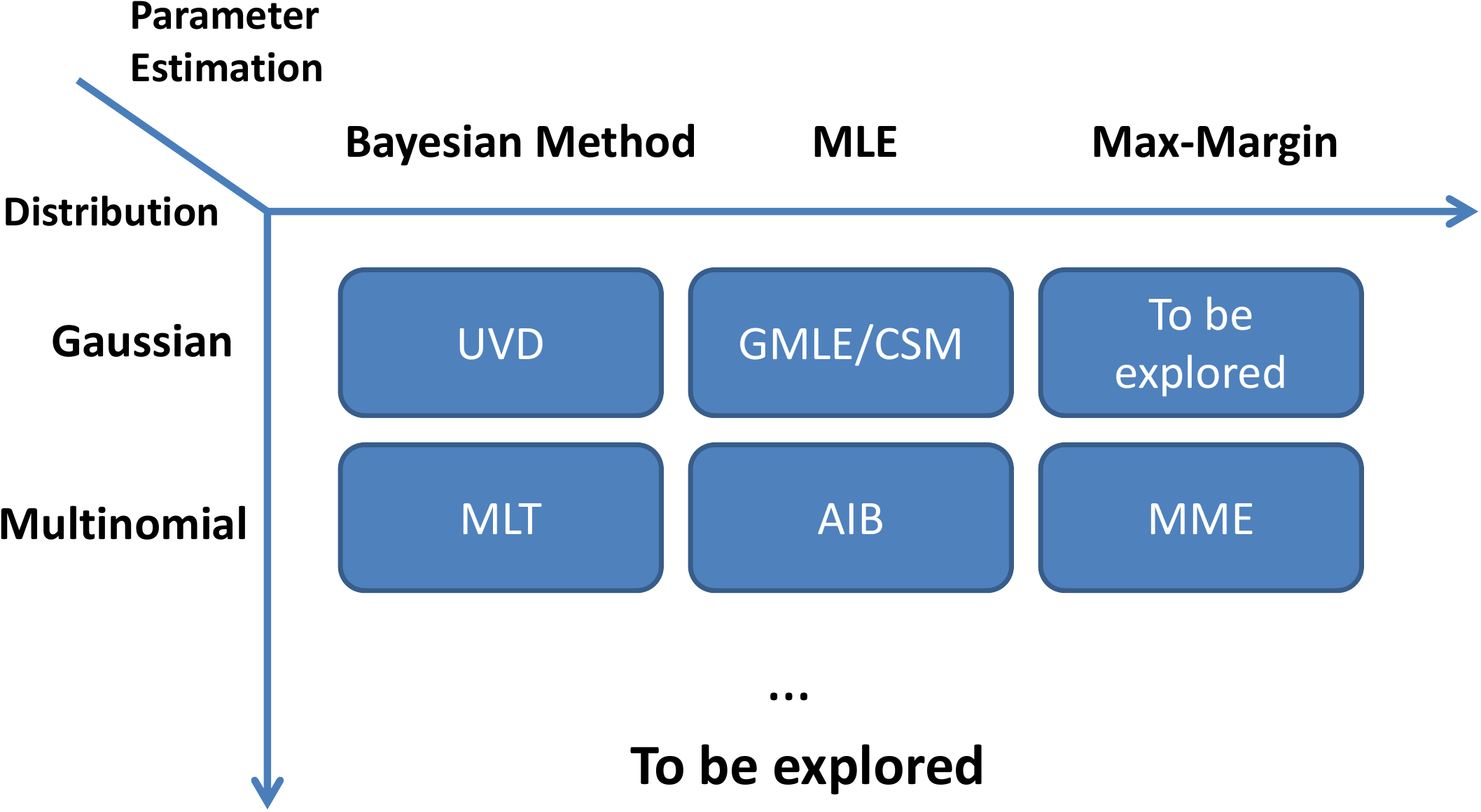}
    \caption{
    Illustration of the proposed framework.
    }
    \label{fig:Framework}
    \end{figure}

  Figure \ref{fig:Framework} is used to illustrate the proposed probabilistic framework. By setting the two factors in different ways, the framework not only accommodates the existing criteria UVD, AIB and CSM, but also produces a matrix of new criteria. 
  Three of them, called MLT, GMLE and MME in short, will be investigated. 
  
  In the following sections, we firstly interpret existing methods from the viewpoint of our framework. More specifically, after a brief interpretation of UVD in Section \ref{subsec:UVD}, we show in Section \ref{subsec:AIB} that AIB is a special case of our framework, which chooses the two factors as multinomial distribution and maximum likelihood estimation; In Section \ref{subsec:CSM}, we show that CSM can be (approximately) interpreted as a special case of our framework, which chooses the two factors as Gaussian distribution and maximum likelihood estimation; Then a discussion about the impact of the two factors is given in Section \ref{subset:discussion}. After that, we propose three new merging criteria from Section~\ref{subsec:MLT} to \ref{subsec:MME}.
  From now on, we define $\mathcal J = \log{P({\mathcal H}^{t}|\hat{\mathbf c})}/{P({\mathcal H}^{t}|{\mathbf c}^{\mathrm{same}})}$ and use it throughout the following sections.

\subsection{UVD~\cite{UVD}: Gaussian distribution +  Bayesian method (gamma distribution prior)}\label{subsec:UVD}
	Our framework is inspired by the formulation of UVD, and therefore UVD naturally fits our framework. It uses Gaussian distribution to model the image representation and employs the Bayesian method for parameter estimation. As mentioned above, UVD does not explicitly estimate the model parameters. Instead, it treats the model parameters as random variables and models their distribution through a prior distribution with a set of hyper-parameters.
	
\subsection{AIB~\cite{smartdictionary}: Multinomial distribution + Maximum Likelihood Estimation}\label{subsec:AIB}
	Multinomial distribution\footnote{Strictly speaking, the case in AIB is not exactly a multinomial distribution, and calling it categorical distribution may be more precise. However, these two terms are usually used equivalently in text analysis and we follow this convention in this paper.} has been widely used in the literature to model the occurrence of words in a document. With multinomial distribution, the conditional probability of a histogram is modelled as
\begin{align}
	P(\mathbf{h}) =  \prod_{j} P(v_j)^{{h}_j}.
\end{align}Assuming the i.i.d. property of samples and plugging this distribution model into our framework, we obtain $P({\mathcal H}|{\mathbf c})$ as
    \begin{eqnarray}\label{AIB_Likelihood}
        P({\mathcal H}|{\mathbf c}) = \prod_{\{i|{\mathbf h}_i\in{{\mathcal D}_c}\}} P({\mathbf h}_{i}|\boldsymbol\theta_{c}) 
         =  \prod_{\{i|{\mathbf h}_i\in{{\mathcal D}_c}\}}\prod_{j=1}^{t} P(v_j|c)^{{h}_{ij}}.
    \end{eqnarray}Thus, the merging criterion becomes:
    \begin{eqnarray}\label{AIB_Criterion}
       \mathcal J & = & \sum_{c=1}^{C}\sum_{j =1}^{t} \bar{h}_{cj} \log P(v_j|c) \nonumber\\
       &&-\sum_{j=1}^{t} \left(\sum_{c=1}^{C}\bar{h}_{cj}\right)\log P(v_j|{\mathbf c}^{\mathrm{same}}) \nonumber \\
       & = & \sum_{c=1}^{C}\sum_{j =1}^{t} \bar{h}_{cj} \log \frac{P(v_j|c)}{P(v_j|{\mathbf c}^{\mathrm{same}})},
    \end{eqnarray}where $\bar{h}_{cj}$ denotes the mean of the $j$th bin in class $c$. With training samples, it is not difficult to obtain the MLE of the model parameters as
    \begin{eqnarray}\label{AIB_MLE}
       & P(v_j|c) =  \frac{\bar{h}_{cj}}{\sum_{j=1}^t \bar{h}_{cj}},~~~~ \nonumber \\
       & P(v_j|{\mathbf c}^{\mathrm{same}}) = \frac{\sum_{c=1}^C \bar{h}_{cj}}{\sum_{j=1}^t \sum_{c=1}^C \bar{h}_{cj} }.
    \end{eqnarray}Note that $P(v_j|{\mathbf c}^{\mathrm{same}}) = P(v_j)$ because all samples are assumed to be in a same class in the ${\mathbf c}^{\mathrm{same}}$ configuration. In AIB \cite{smartdictionary}, the terms of $P(v_j|c)$ and $P(v_j)$ are computed in the same way as in Eq.(\ref{AIB_MLE}) \footnote{This can be seen in the code provided in \cite{vlfeat}.}. Also, AIB computes the joint probability as
    \begin{eqnarray}\label{AIB_MLE2}
        P(v_j,c)  = \frac{\bar{h}_{cj}}{\sum_{j=1}^t \sum_{c=1}^C \bar{h}_{cj} }.
    \end{eqnarray}Note that the denominator $\sum_{j=1}^t \sum_{c=1}^C \bar{h}_{cj} $ keeps constant when merging different words at the level $t$. Substituting $\bar{h}_{cj} = P(v_j,c)\sum_{j=1}^t \sum_{c=1}^C \bar{h}_{cj} $ into Eq.(\ref{AIB_Criterion}) and dropping constant $\sum_{j=1}^t \sum_{c=1}^C \bar{h}_{cj}$, we produce AIB criterion in \cite{smartdictionary} because
    \begin{equation}\label{AIB_Criterion-1}
       {\mathrm{Eq.}(9)} \propto \sum_{c=1}^{C}\sum_{j =1}^{t} P(v_j,c) \log \frac{P(v_j,c)}{P(v_j)P(c)} = {\mathrm{AIB}}.
    \end{equation}

\subsection{CSM~\cite{Fast2008Wang}: Gaussian distribution + Maximum Likelihood Estimation}\label{subsec:CSM}

By modelling training data with a Gaussian distribution, Eq. (\ref{FrameworkP}) will lead to a criterion shown below.
     \begin{eqnarray}\label{Gaussian_Likelihood}
        & & P({\mathcal H}|{\mathbf c}) =  \prod_{\{i|{\mathbf h}_i\in{{\mathcal D}_c}\}} P({\mathbf h}_{i}|\boldsymbol\theta_{c}) \nonumber \\  
        & \propto |{\boldsymbol  \Sigma }_{c}|^{-\frac{N_c}{2}} & \exp\left(-\frac{1}{2} \sum_{\{i|{\mathbf h}_i\in{{\mathcal D}_c}\}} ({\mathbf h}_{i} - {\boldsymbol\mu}_c)^{\top} {\boldsymbol  \Sigma}_{c}^{-1}({\mathbf h}_{i} - {\boldsymbol\mu}_c)\right), \nonumber\\
    \end{eqnarray}where ${\boldsymbol\mu}_c$ and$\ {\boldsymbol\Sigma}_{c}$ denote the mean and the covariance matrix for class $c$. $N_c$ is the number of training samples in class $c$. Then $\mathcal J = \log{P({\mathcal H}|\hat{\mathbf c})}/{P({\mathcal H}|{\mathbf c}^{\mathrm{same}})}$ becomes
     \begin{eqnarray}\label{Gaussian_Criterion}
        \mathcal J & = &  \mathrm{const.} + \log\frac{|\boldsymbol\Sigma_c|}{|\boldsymbol\Sigma|} + \sum_{i=1}^{n}({\mathbf h}_{i} - {\boldsymbol\mu})^{\top} {\boldsymbol\Sigma}^{-1}({\mathbf h}_{i} - {\boldsymbol\mu}) \nonumber\\
        &  & -\sum_{c = 1}^{C} \sum_{\{i|{\mathbf h}_i\in{{\mathcal D}_c}\}} ({\mathbf h}_{i} - {\boldsymbol\mu}_c)^{\top} {\boldsymbol  \Sigma}_{c}^{-1}({\mathbf h}_{i} - {\boldsymbol\mu}_c),
    \end{eqnarray}where ${\boldsymbol\mu}$ and$\ {\boldsymbol\Sigma}$ denote the total mean and the covariance matrix for all data. Assuming that ${\boldsymbol\Sigma}_{1} = {\boldsymbol\Sigma}_{2} = ...= {\boldsymbol\Sigma}_{C} = {\mathrm{diag}}(\sigma^2_1,..,\sigma^2_1)$ and ${\boldsymbol\Sigma} = {\mathrm{diag}}(\sigma^2_0,..,\sigma^2_0)$, Eq.(\ref{Gaussian_Criterion}) can be simplified as 
    \begin{eqnarray}\label{TraceminusTrace}
        \mathcal J & = & \mathrm{const.} + \frac{1}{\sigma^2_0} \sum_{i=1}^{n}\| {\mathbf h}_{i}  - {\boldsymbol\mu}\|^{2} - \nonumber \\
        && \frac{1}{\sigma^2_1} \sum_{c = 1}^{C} \sum_{\{i|{\mathbf h}_i\in{{\mathcal D}_c}\}} \| {\mathbf h}_{i} - {\boldsymbol\mu}_c\|^{2}   \nonumber \\
        & \propto & - \left({\mathrm{tr}}({\mathbf S}_{w}) - (\sigma^2_1/\sigma^2_0) {\mathrm{tr}}({\mathbf S}_{t}) \right),
    \end{eqnarray}where ${\mathbf S}_{w}$ and ${\mathbf S}_{t}$ are the within-class scatter matrix and the total scatter matrix defined in \cite{Fast2008Wang}. The criterion ${\mathrm{tr}}({\mathbf S}_{w}) - (\sigma^2_1/\sigma^2_0){\mathrm{tr}}({\mathbf S}_{t})$ strongly connects with ${\mathrm{tr}({\mathbf S}_{w})}/{\mathrm{tr}({\mathbf S}_{t})}$ used in \cite{Fast2008Wang}. Minimizing ${\mathrm{tr}({\mathbf S}_{w})}/{\mathrm{tr}({\mathbf S}_{t})}$ is a fractional programming problem. It can be effectively solved by the Dinkelbach's algorithm~\cite{SDP2008Shen}, which iteratively minimizes ${\mathrm{tr}}({\mathbf S}_{w}) - \lambda{\mathrm{tr}}({\mathbf S}_{t})$, where $\lambda$ is the ratio of ${\mathrm{tr}}({\mathbf S}_{w})$ to ${\mathrm{tr}}({\mathbf S}_{t})$ at the last iteration.

\subsection{Discussion on the two key factors}\label{subset:discussion}
\textit{Parameter estimation.} In UVD, parameter estimation is implicitly handled through the Bayesian method. The performance of the Bayesian method highly depends on the choice of prior distribution and its hyper-parameters. In practice, for the sake of computational feasibility, the hyper-parameters are usually empirically set and a same set of hyper-parameters is often applied to all classes. This could bring negative impact to the practical performance of the Bayesian method. As a result, the Bayesian method does not necessarily outperform the way that explicitly estimates model parameters from training data, for example, through maximum likelihood estimate~(MLE).

\textbf{Distribution model:} If the true distribution of data is known, we could employ it in our framework and produce a merging criterion of high quality. In practice, however, we do not have such information and have to rely on our knowledge to choose the distribution model. The appropriateness of the chosen model plays a pivotal role.

From the three existing merging criteria discussed above, two distributions are employed, namely, multinomial distribution and Gaussian distribution. Note that in the literature, the BoW model originates from document analysis, in which a histogram of words is usually modelled by a multinomial distribution~\cite{LDA}. In this sense, multinomial distribution seems to be a suitable choice of modelling histogram based image representation. On the other hand, with the recent development of bag-of-features model, the local features are usually sampled at a dense spatial grid \cite{DBLP:conf/iccv/JurieT05}. This operation in effect reduces the sparsity of the histogram and may change the underlying distribution of training data. In addition, some post-processing such as square root operation \cite{UVD} on the histogram could also alter the characteristics of the distribution of training data\footnote{For example, as indicated in UVD \cite{UVD}, square rooting operation has the effect of making the data distribution to be more Gaussian-alike.}. In this work, we find that Gaussian distribution sometimes results in a good merging criterion too when the image representation is obtained by using the dense sampling and square root operation. Note that in our previous study on this framework \cite{ProbFramework11}, these settings have not been considered.

\subsection{MLT: Multinomial distribution + Bayesian Method (Dirichlet prior)}\label{subsec:MLT}

    In this section, we first propose to use the multinomial distribution and Dirichlet prior to replace the Gaussian distribution and the Gaussian-gamma prior in UVD~\cite{UVD}. This will produce a new merging criterion called MLT. This new criterion can outperform UVD when data is better characterized by multinomial distribution.

    In MLT, $\ P({\mathcal H}|{\mathbf c})$ is still modeled as Eq.(\ref{eqn:BayesianModel}), but the likelihood and the prior terms become:
    \begin{eqnarray}\label{UVDDModel}
    P({\mathbf h}_{i}|\boldsymbol\theta_{c}) &=& \prod_{j=1}^{t} P(v_j|c)^{{h}_{ij}} \nonumber \\     
    P(\boldsymbol\theta_{c}) &=& \frac{1}{B(\boldsymbol\alpha)}\prod_{j=1}^{t} P(v_j|c)^{\alpha_{j}-1},
    \end{eqnarray}where $v_j$ denotes the $j$th word and $P(v_j|c)$ is the model parameter, which represents the likelihood of word $v_j$ occurring in class $c$. $B(\boldsymbol\alpha)$ is the multinomial Beta function and $\boldsymbol\alpha = (\alpha_{1},...,\alpha_{t})$ is the hyper-parameter. Substituting Eq.(\ref{UVDDModel}) into Eq.(\ref{eqn:BayesianModel}), we can derive that

    \begin{eqnarray}\label{UVDD_likelihood}
        P({\mathcal H}|{\mathbf c}) = \cdots = \prod_{c=1}^{C} \frac{B(\boldsymbol\alpha + \bar{\mathbf h}_{c})}{B(\boldsymbol\alpha)},   
    \end{eqnarray}where we define $\bar{\mathbf h}_{c} = (\bar{h}_{c1},...,\bar{h}_{ct})$ for class $c$ and $ \bar{h}_{cj} = \sum_{\{i|{\mathbf h}_i\in{{\mathcal D}_c}\}}{h}_{ij}$.
    
    Note that the integral in Eq. (\ref{eqn:BayesianModel}) can be analytically worked out in this case because the Dirichlet distribution is the conjugate prior of a multinomial distribution. In this way, the proposed MLT criterion is obtained as
    \begin{eqnarray}\label{UVDD_criterion1}    
    \mathcal J = \sum_{c=1}^{C}\log B(\boldsymbol\alpha + \bar{\mathbf h}_{c}) - \log B(\boldsymbol\alpha + \sum_{c = 1}^{C} \bar{\mathbf h}_{c}) + const.
    \end{eqnarray}Recall that $\mathcal J = \log{P({\mathcal H}|\hat{\mathbf c})}/{P({\mathcal H}|{\mathbf c}^{\mathrm{same}})}$. At each level of the hierarchy, the pair of words $r$ and $s$ whose mergence maximizes $\mathcal J$ is identified and merged.

\subsection{GMLE: Gaussian distribution + Maximum Likelihood Estimation}
From Section~\ref{subsec:CSM}, we can see that CSM~\cite{Fast2008Wang} simply treats the covariance matrices as a scaled identity matrix and uses a formulation that only approximately connects with the proposed framework, as shown in Eq.(\ref{TraceminusTrace}). In this section, we propose another merging criterion by following the proposed framework, which is called GMLE in short.

For the sake of reliable parameter estimation and the computational efficiency of GMLE, we assume that the covariance matrices are diagonal (but not an identity matrix), that is, $\boldsymbol\Sigma_{c} = \mathrm{diag}(\sigma^2_{c1},..,\sigma^2_{ct})$ and $\boldsymbol\Sigma = \mathrm{diag}(\sigma^2_{1},..,\sigma^2_{t})$. By doing so, we can rewrite the criterion in Eq. (\ref{Gaussian_Criterion}) as:
\begin{eqnarray}\label{Gaussian_Criterion2}
 \mathcal J & = & \mathrm{const.} + \sum_{j=1}^{t} \log(\frac{\sigma_{cj}}{\sigma_{j}}) + \sum_{j=1}^t \sum_{i=1}^{n}({h}_{ij} - {\mu}_j)^2 \frac{1}{\sigma_{j}^2} \nonumber\\
        &  & -\sum_{j=1}^{t}\sum_{c = 1}^{C} \sum_{i=1}^{n}({h}_{cij} - {\mu}_{cj})^2 \frac{1}{\sigma_{cj}^2}.
    \end{eqnarray}
Note that $\mathcal J$ is a summation over terms depending on each dimension. For a given merging pair $r$ and $s$, the criterion value can be efficiently re-evaluated by only updating the terms involving $r$ and $s$, which significantly reduces the computational cost. 

\subsection{MME: Multinomial distribution + Max-Margin Parameter Estimation}\label{subsec:MME}

    The maximum likelihood estimation (MLE) of model parameters still presents potential drawbacks. Due to its generative nature, it prevents us from using more information in training data. Particularly, when the multinomial distribution is employed, the MLE of its parameters are only determined by the average histogram per class and higher order statistics such as the variances of visual words are completely neglected. Thus the creation of a compact codebook does not fully exploit the information of training data and consequently the resulted performance may be less satisfying. In the literature, this phenomenon is known as \textit{exchangeable property} \cite{LDA}. One way to overcome this drawback is to adopt more complex distributions, for example, the multivariate Polya distribution \cite{Multiploya}. However, this will lead to intractable computation because there is usually no analytical MLE for the parameters in these complex models. Another disadvantage of MLE is that the estimation could become unreliable when training samples are scarce or many less discriminative visual words exist. MLE estimates cannot effectively identify the discriminative words since the parameters are estimated based on the data from each class individually. This limits the performance of the created compact codebooks.

    To improve this situation, we propose a new Max-Margin parameter Estimation (MME) scheme for merging visual words. The idea is to seek the model parameters that can \textit{maximize the margin of posterior probability ratio of the true class label to all other possible labels} under certain regularization. The disadvantages of MLE mentioned above can be removed because (i) the parameter estimation now considers all training samples from different classes together; (ii) the max-margin principle emphasizes discriminative features. In the remaining parts of this section, we firstly present a detailed derivation of the proposed max-margin parameter estimation formulation. Then we discuss how to solve the resultant optimisation problem and the implementation.

\subsubsection{Problem formulation}\label{sect:problem formulation}
We still model $P({\mathbf h}_{i}|\boldsymbol\theta_{c})$ by a multinomial distribution. The posterior probability ratio for the $i$-th training sample is defined as
    \begin{eqnarray}\label{Posterior_Ratio}
        & & R_{i,c} = \log\frac{P(c_i|{\mathbf h}_{i})}{P(c|{\mathbf h}_{i})} = \log\frac{P({\mathbf h}_{i}|c_i)P(c_i)}{P({\mathbf h}_{i}|c)P(c)} \nonumber \\
        & = & \sum_{j = 1}^{t} h_{ij} \log\frac{P(v_j|c_i)}{P(v_j|c)} + \log\frac{P(c_i)}{P(c)}, ~\forall{c} \neq c_i;
    \end{eqnarray}where $c_i$ is the true label of sample $i$ and $c$ is one of the other possible labels. Note that this ratio will take a form of linear classifier if we treat $\log\frac{P(v_j|c_i)}{P(v_j|c)}$ and $\log\frac{P(c_i)}{P(c)}$ as variables, although the parameters to estimate are $P(v_j|c_k)$ and $P(c_k),~~k = 1,..,C$ and $j = 1,..,t$. This ratio reflects how confident the sample $i$ is classified into its ground-truth class $c_i$ and a large ratio is preferred. 

	The idea of max-margin parameter estimation can be intuitively understood as to maximize the lowest $R_{i,c}$ for all pairs of $i$ and $c$, that is, to maximize the minimum confident score. However, merely optimizing $\min_{i=1,\cdots,n \atop c=1,\cdots,C}R_{i,c}$ could lead to severe over-fitting because the ratio $R_{i,c}$ can always be increased by reducing $P(v_j|c)$ towards zero. To avoid such a situation, we introduce a regularization term 
\begin{align}
\Upsilon  = \sum_{p,q}\left[\sum_{j=1}^{t} \left(\log\frac{P(v_j|c=p)}{P(v_j|c=q)}\right)^2 + \alpha \left(\log\frac{P(c=p)}{P(c=q)}\right)^2\right],
\end{align}where $\alpha$ is a positive constant which controls the relative strength of the regularization on $P(v_j|c)$ and $P(c)$. This term attains its minimum  when $P(v_j|c=p) = P(v_j|c=q)$ and $P(c=p) = P(c=q)$. Thus, it prefers a uniform estimation of $P(v_j|c)$ and $P(c)$ with respect to different class $c$, which is consistent with the principle of maximum entropy \cite{MaxEnt}. Inspired by the margin definition in SVM \cite{BayesianPointMachine}, we formally define the margin of the posterior probability ratio as the ratio between the minimal $R_{i,c}$ and the regularization term:
\begin{align} \label{eqn:max-margin1}
        &\rho  =  \min_{i=1,\cdots,n \atop c=1,\cdots,C}\frac{R_{i,c}}{\sqrt{\Upsilon}} =  \frac{\min_{i=1,\cdots,n \atop c=1,\cdots,C} R_{i,c}}{\sqrt{\Upsilon}} \nonumber \\
        & \forall ~ p\neq{q}, ~~p,q = 1,2,\cdots,C.
\end{align}Then the proposed max-margin parameter estimation aims to maximize $\rho$ over the model parameters. Note that scaling the terms $\log\frac{P(v_j|c=p)}{P(v_j|c=q)}$ and $\frac{P(c=p)}{P(c=q)}$ does not change the value of $\rho$. As a result, the solution of this margin maximization problem is not unique and different solutions are connected through a scaling factor. Without loss of generality, we can obtain one of these solutions by simply setting the regularization term to be a positive constant $\epsilon$, that is,

\begin{align}\label{eqn:additional-constraint}
\sum_{p,q}\left[\sum_{j=1}^{t} \left(\log\frac{P(v_j|c=p)}{P(v_j|c=q)}\right)^2 + \alpha \left(\log\frac{P(c=p)}{P(c=q)}\right)^2\right] = \epsilon.
\end{align} 
When $\epsilon$ is fixed, the margin maximization problem becomes:
   \begin{align} \label{max-margin1}
        & \max  \quad \gamma \nonumber\\
        s.t \quad & \sum_{j = 1}^{t} h_{ij} \log\frac{P(v_j|c_i)}{P(v_j|c)} + \log\frac{P(c_i)}{P(c)}  \geq \gamma ,~~ \forall ~ c \neq c_i;  \nonumber\\
        &  \sum_{p,q}\left[ \sum_{j=1}^{t} \left(\log\frac{P(v_j|c=p)}{P(v_j|c=q)}\right)^2 + \alpha \left(\log\frac{P(c=p)}{P(c=q)}\right)^2 \right] = \epsilon; \nonumber \\
        & \forall ~ i = 1,2,\cdots,n; \nonumber \\
        & \forall ~ p\neq{q}, ~~p,q = 1,2,\cdots,C.
    \end{align}Let us define $w_{pq}^j= \log\frac{P(v_j|c=p)}{P(v_j|c=q)}$ and $b_{pq}= \log\frac{P(c=p)}{P(c=q)}$. By re-scaling $w_{pq}^j$ and $b_{pq}$ by $\frac{1}{\gamma}$ ($\gamma > 0$ for linear separable case), it is easy to rewrite the above maximisation problem as
     \begin{align} \label{max-margin3}
        & \min_{w_{pq},b_{pq}}   \quad \frac{\epsilon}{\gamma^2} \nonumber\\
        s.t \quad & \sum_{j = 1}^{t} h_{ij} \frac{w_{c_ic}^j}{\gamma} + \frac{b_{c_ic}}{\gamma}  \geq  1 ,~~ \forall ~ c \neq c_i;  \nonumber\\
        &  \sum_{\forall ~ p\neq{q}}\left[ \sum_{j=1}^{t} \left( \frac{w_{pq}^j}{\gamma} \right)^2 + \alpha \left( \frac{b_{pq}}{\gamma} \right)^2 \right] = \frac{\epsilon}{\gamma^2}; \nonumber \\
        & \forall ~ i = 1,2,\cdots,n; \nonumber \\
    \end{align}Furthermore, by defining $\bar{w}_{pq}^j \triangleq \frac{w_{pq}^j}{\gamma}$ and $\bar{b}_{pq} \triangleq \frac{b_{pq}}{\gamma}$, Eq. (\ref{max-margin3}) can be expressed in a compact form as
    \begin{align} \label{max-margin4}
        & \min_{\bar{w}_{pq},\bar{b}_{pq}}   \quad \sum_{\forall ~ p \neq q}\left[ \sum_{j=1}^{t} \left( \bar{w}_{pq}^j \right)^2 + \alpha \left( \bar{b}_{pq} \right)^2 \right] \nonumber\\
        s.t \quad & \sum_{j = 1}^{t} h_{ij} \bar{w}_{c_ic}^j + \bar{b}_{c_ic}  \geq  1 ,~~ \forall ~ c \neq c_i;  \nonumber\\
        & \forall ~ i = 1,2,\cdots,n; \nonumber \\
    \end{align} 
    It is worth mentioning that if our interest is to learn a model for classification, we can simply treat $\bar{w}_{pq}^j$ and $\bar{b}_{pq}$ as variables and the scaling of $\bar{w}_{pq}^j$ and $\bar{b}_{pq}$ will not change the decision function. This is why the scaling factor $\frac{1}{\gamma}$ is usually ignored in max-margin learning problems, e.g. SVMs. However, our goal is to estimate $P(c)$ and $P(v_j|c)$, for which the scaling factor will affect the estimation. Hence, $\frac{1}{\gamma}$ has to be explicitly considered in our case.

\subsubsection{Problem solution and implementation}\label{sect:MME_solution}

It is not difficult to see that the problem in Eq.(\ref{max-margin4}) is similar to a linear SVMs. In fact, if we consider binary classification (which is the focus of this paper) and add the slack variables to handle the non-separable case, Eq.(\ref{max-margin4}) will reduce to a standard binary linear SVM with several additional constraints, that is:
\begin{align} \label{final_problem}
        & \min_{w_j,b,P(v_j|c),P(c),\xi_{i}}  \sum_{j=1}^{t} w_j^2 + \alpha b^2 +  \lambda \sum_{i}\xi_{i}  \nonumber\\
        s.t \quad & \sum_{j = 1}^{t} h_{ij} w_j + b  \geq  1 - \xi_{i} , \nonumber\\
			\quad & \xi_{i} \ge 0 ~~ \forall ~i; \nonumber\\
			\quad & \eta w_j = \log\frac{P(v_j|c = 1)}{P(v_j|c = -1)},\nonumber\\
			\quad & \eta b = \log \frac{P(c = 1)}{P(c = -1)}; ~~\eta > 0 \nonumber \\
			\quad & \sum_{c=-1,+1}P(v_j|c)P(c) = P(v_j),~~ \sum_{c=-1,+1} P(c) = 1. \nonumber \\
			\quad & 0 \le P(v_j|c) \le 1, 0 \le P(c) \le 1, ~~\forall ~j,c,
\end{align}where we define $w_j = w_{1,-1}^j, b = b_{1,-1}, \eta = \frac{1}{\gamma}$ for the symbol simplicity. The first two constraints are identical to those in the standard SVM. The third and fourth constraints establish the relationship between the SVM solution and the multinomial distribution parameters, where $\eta$ is the scaling factor discussed in subsection \ref{sect:problem formulation}. The last two constraints come from the properties of probability. Note that we need to incorporate $P(v_j)$ to make the variables properly bounded. In this work, we simply obtain $P(v_j)$ via the method of MLE.

At the first glance, solving this optimization problem is hard since it involves many nonlinear constraints. However, we show that under mild assumptions, the problem in Eq.(\ref{final_problem}) could be solved in two stages. At the first stage, we only consider the first two constraints to construct a sub-problem and obtain the solution via an off-the-shelf SVM solver. At the second stage, we calculate the probability parameters by solving equations $\sum_{c=-1,+1}P(v_j|c)P(c) = P(v_j)$ and $\sum_{c=-1,+1} P(c) = 1 ~~\forall~i,c$. The key insight here is that if a certain assumption is taken for a given solution of $\{w_j\}$ and $b$, we can always find corresponding values for $P(v_j|c)$ and $P(c)$ to make the remaining constraints satisfied. The detailed analysis is presented in Appendix A. 

After obtaining these model parameters, we can readily apply them to the multinomial distribution to compute $\mathcal{J}$ to identify the optimal pair of words to merge at each level of $t$.

\noindent \textbf{Implementation}

\begin{itemize}
\item{
To evaluate $\mathcal{J}$ for a pair of words $r$ and $s$, we need to calculate the class conditional probability $P(v_{rs}|c)$ for the merged word $v_{rs}$. If strictly following the max-margin parameter estimation, we have to re-estimate $P(v_{rs}|c)$ by solving Eq.(\ref{final_problem}) for each possible pair of $r$ and $s$, incurring a computational cost at the order of ${\mathcal O}(t^2)$ at level $t$. Even though a highly efficeint SVM solver is used, this repeated re-estimation process will still be too time-consuming. In practice, we adopt a compromised scheme: the max-margin estimation is only carried out once at each level after the optimal pair of words is identified. In the course of identifying the optimal pair, the updating formula$\ P(v_{rs}|c) = P(v_r|c) + P(v_s|c)$ is used for the merged word $v_{rs}$. That is, the underlying criterion for identifying the optimal merging pair at each level $t$ is same to that used in AIB, while the model parameters are estimated via the proposed max-margin estimation scheme. Experimental study shows that this strategy works very well in practice.
}
\item{In our implementation, we use LIBSVM to solve Eq. (\ref{final_problem}). We set $\alpha=0$ in order to be consistent with the formulation used in LIBSVM. Also, since LIBSVM solves SVM in its dual form, we use the precomputed kernel as the input of LIBSVM interface. At each hierarchy, two words $r$ and $s$ are identified and merged. This leads to an update of kernel matrix which could be efficiently calculated by
\begin{eqnarray}\label{Gaussian_Criterion2}
 	\mathbf{K}_{t-1} = \mathbf{K}_{t} + \mathbf{h}_r^t {\mathbf{h}_s^t}^T + \mathbf{h}_s^t {\mathbf{h}_r^t}^T,
\end{eqnarray}  
where $\mathbf{K}_{t}$ is the kernel matrix at the $t$th level. $\mathbf{h}_r^t = {(h_{1r}^t, h_{2r}^t \cdots, h_{tr}^t)}^T$. $\mathbf{h}_s^t$ is defined in a similar way. Note that this update is efficient since  $\mathbf{h}_r^t$ and $\mathbf{h}_s^t$ are merely two column vectors.
}

\end{itemize}

\section{Experimental Result}

To examine the effectiveness of our framework and the impact of the two factors identified in our framework, we conduct a number of experiments in this section. In our experiment, the goodness of a compact codebook is evaluated by its performance on two applications: 1) Building compact representation for image classification. In this application, the aim is to create a compact image representation which can largely maintain the discriminative power of the initial codebook. The performance of a word merging method is evaluated by the classification performance with respect to the reduced codebook size. 2) Using compact codebook for efficient pixel-level object detection. This is an application in which the use of compact codebook could significantly reduce the computational complexity. The aim of using this application for evaluation is to see whether the newly proposed methods can achieve better performance than the traditional ones in a real-world application.

The experiments are organized into two parts. The first part is based on the first application and the purpose of this part is to demonstrate the impact of the two key factors identified in our framework. More specifically, we conduct three experiments in this part.
\begin{itemize}
 \item (1) The evaluation of MLT. In this experiment, we focus on the comparison between MLT and UVD. This comparison aims to show the importance of choosing appropriate distribution model in our framework. 
 \item (2) The evaluation of GMLE. This experiment focuses on the comparison between GMLE and UVD. This comparison aims to validate the use of MLE as an appropriate parameter estimation method in our framework.
 \item (3) The evaluation of MME. The purpose of this experiment is to demonstrate the advantage of using max-margin parameter estimation in our framework.
\end{itemize}

In the second part of our experiments, we further show the excellent performance of the proposed methods, especially MME on the second application. 

In the proposed MME method, there is a scaling factor $\eta$ which can be chosen freely within a range. To investigate its impact on the performance of MME, we also conduct theoretical and experimental analysis on the choice of its value.

Throughout the experiments, six methods induced from our framework are compared. They are AIB \cite{smartdictionary}, UVD \cite{UVD}, CSM \cite{Fast2008Wang}, MLT, GMLE and MME. Also, we focus on the binary-class classification/detection setting. Multi-class case can be handled by one-vs-rest decomposition.

Five datasets are used in our experiments, including Caltech-256 \cite{Caltech256}, PASCAL VOC2007, PASCAL VOC2012 \cite{PASCAL}, KTH \cite{KTH} and Graz-02 \cite{Graz02}. The first four datasets are used for the evaluation of image-level classification task while the last one is mainly used for the evaluation of pixel-level object detection task. The introduction of these datasets and their preprocessing details are elaborated as follows:

	\textit{(1) Caltech-256} Caltech-256 contains 256 object classes and one background class. For this dataset, we create 256 object-vs-background classification tasks, that is, the task is to discriminate the images containing the object from the background class images. For each object-vs-background task, we randomly split the images into 10 training/test sets and report the average performance of all 10 splits. To obtain the bag-of-features image representation, we firstly densely sample $16\times16$ patches with the step size of 8 pixels and describe them by the SIFT descriptor using the implementation in \cite{SPM}. Then an initial codebook with 1024 visual words is created by applying a $k$-means clustering on the local features sampled from the training images. Finally, we use this codebook to create a histogram for each image. We normalize each histogram to make its $l$-1 norm equal to 1 to eliminate the affect of the image size difference. In our evaluation, we also apply a square-root operation on each histogram since it usually significantly boosts the classification performance. A linear SVM is applied as the classifier and we use LIBSVM \cite{libsvm} as the SVM solver. 

	\textit{(2) PASCAL VOC2007} PASCAL VOC2007 is a commonly used evaluation benchmark for image classification. It contains 20 object classes and in the standard evaluation protocol the task is to distinguish the images containing the object from those that do not. In our experiment, we follow this evaluation protocol and use the training/validation/test sets provided by this dataset. We learn a linear SVM classifier from the training set together with the validation set and evaluate the performance by mean average-precision (mAP) on the test set. We use the same image representation extraction approach as the one used for Caltech-256 but with a larger-sized codebook containing 4000 visual words. 
	
	\textit{(3) PASCAL VOC2012} PASCAL VOC2012 is the latest PASCAL VOC dataset. Except for more images, the other settings and evaluation protocol used for this dataset are identical to those used in PASCAL VOC2007. Since the test set has not been released, we use the validation set as the test set instead. 
	
	\textit{(4) KTH} KTH is a commonly used action recognition benchmark. It consists of six actions: boxing, hand-clapping, jogging, running, walking and hand-waving. These actions are performed by 25 subjects in various scenarios, e.g. different lighting conditions, clothes and viewpoints. In our experiment, we randomly choose the actions performed by 16 subjects as the training set and the actions performed by the remaining 9 subjects as test set. We repeat this random partition ten times and report the average performance on the ten groups of training/test sets. The histogram-of-optical-flow and histogram-of-gradient are extracted as local features at the interest points detected by the spatial-temporal interest point detector \cite{Laptev08}. Following \cite{Laptev08}, we create a 4000-visual-word codebook and represent each video by a histogram of 4000 visual words. Six one-vs-rest classification tasks are used for the evaluation.

	\textit{(4) Graz-02} Graz02 \cite{Graz02} contains three object categories, \textit{Car}, \textit{Person} and \textit{Bike}. The pixel-wise object annotations are provided by this dataset. The task is to learn a detector to determine whether a pixel belongs to the foreground (object) or the background. We follow the framework in \cite{smartdictionary} to use the bag-of-features model to extract the feature representation for each pixel, that is, for each pixel in an image we crop a local region around it and use the histogram of visual words in this region as the feature representation for that pixel. Then this feature is sent to a classifier to test if the pixel belongs to foreground or background. We densely extract SIFT features with the same settings as in the previous experiments and a 1000-word-codebook is utilized. The local region size and the boundary issue are set and handled in the same way as in \cite{smartdictionary}. Following the setting in \cite{smartdictionary} we use the first 150 odd-numbered images as the training set and the first 150 even-numbered images as the test set. The normalization and square root operation are applied to the extracted histograms since they lead to better detection performance. 

	To generate the training set, we randomly sample pixels in a training image and use their feature representations as training samples. The class label of each sample is determined by whether its corresponding pixel belongs to foreground or background. Note that the above procedure is different from the way of generating the training set in \cite{smartdictionary}. In \cite{smartdictionary}, each positive (negative) class sample is the histogram of visual words which are obtained over the whole foreground (background) region in an image rather than the local region centered at each pixel. Compared with their method, our scheme keeps the consistency in sample generation process between the training and test stages. Empirically, we find this simple modification could lead to significant improvement on the detection performance.

\subsection{Comparison of UVD and MLT}

  \begin{figure*}[!ht]
    \begin{tabular}{l}
            \subfloat[]{ \includegraphics[height=43mm,width=50mm]{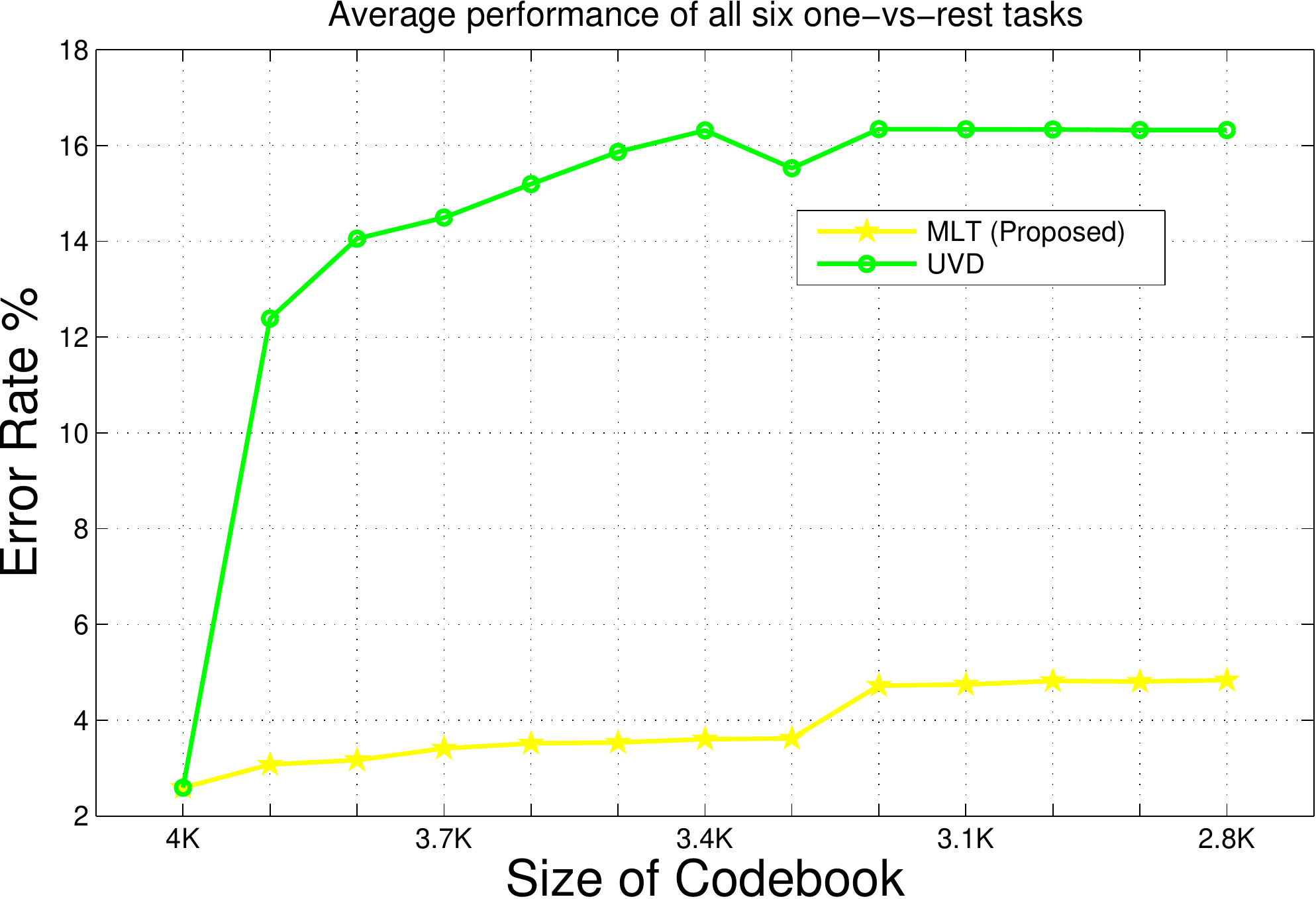}}
            \subfloat[]{ \includegraphics[height=43mm,width=50mm]{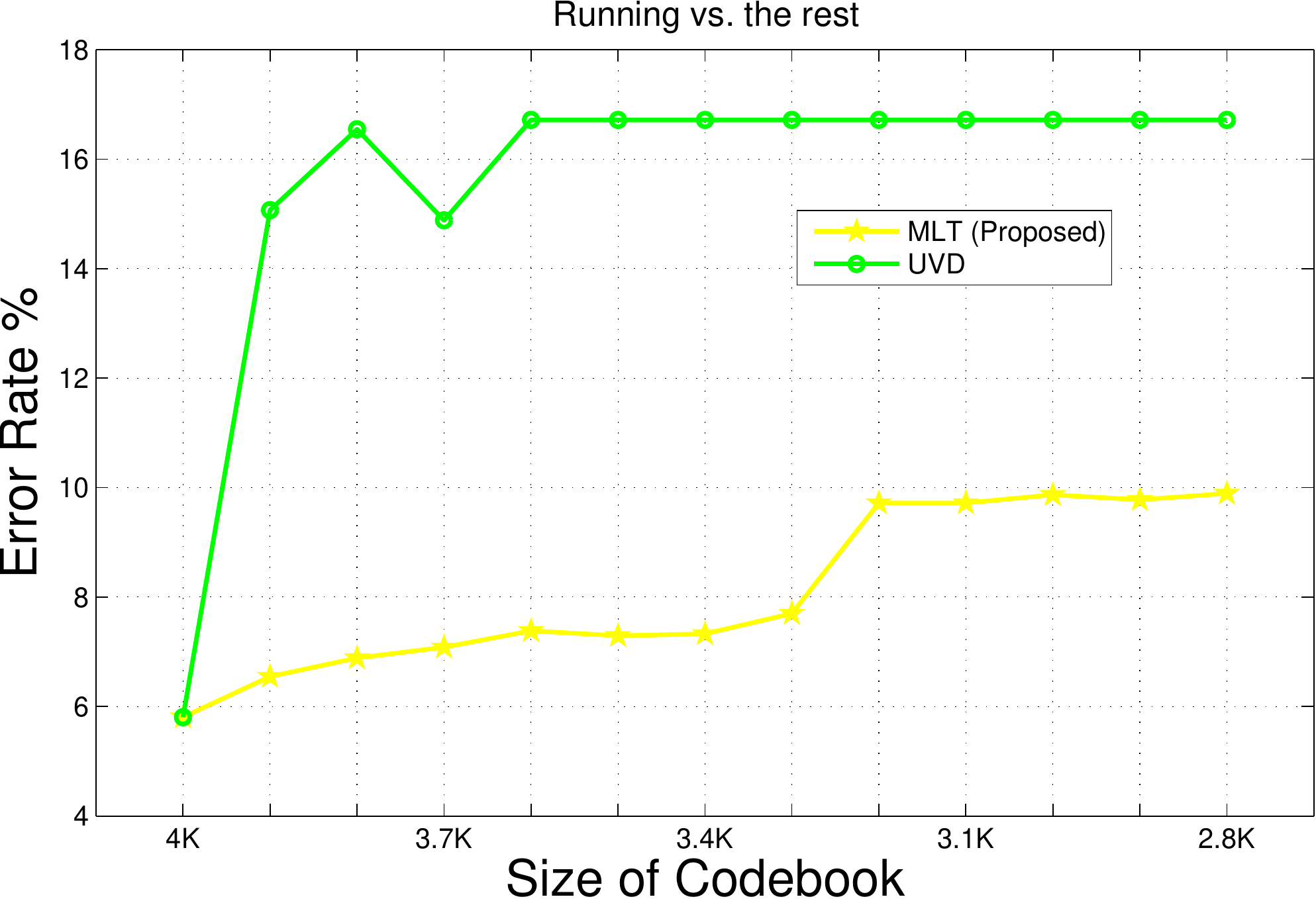}} 
            \subfloat[]{ \includegraphics[height=43mm,width=50mm]{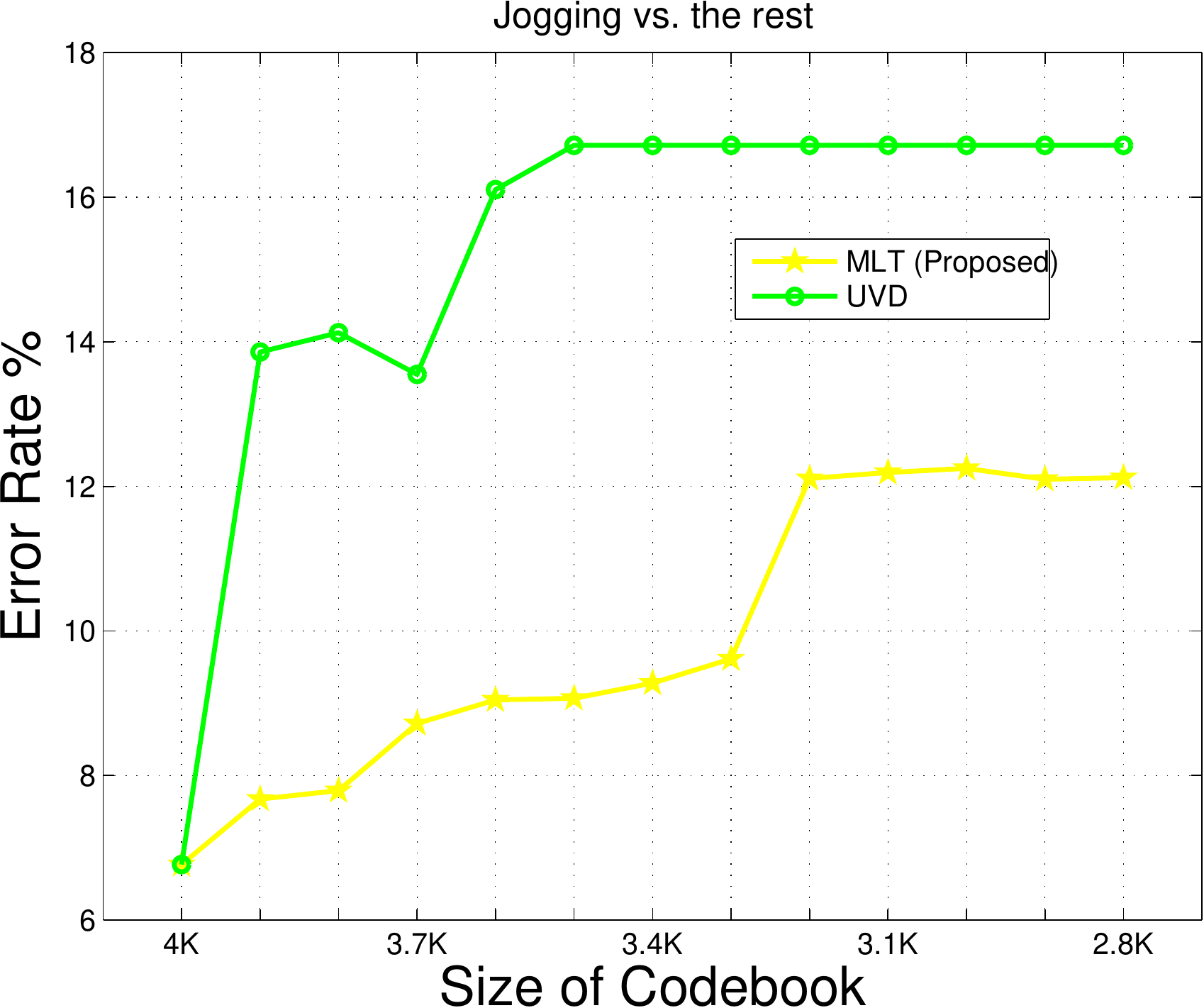}}

    \end{tabular}

    \caption{ Comparison of UVD and MLT on KTH action recognition dataset. Six one-vs-rest tasks are tested. The average performance of six tasks is shown in (a). Two most difficult tasks, ``running vs. the rest'' and ``jogging vs. the rest'' are shown in (b) and (c) respectively.
    }
    \label{fig:KTH_Comparison}
    \end{figure*}

  \begin{figure*}[!ht]
    \begin{tabular}{l}
            \subfloat[]{ \includegraphics[height=43mm,width=50mm]{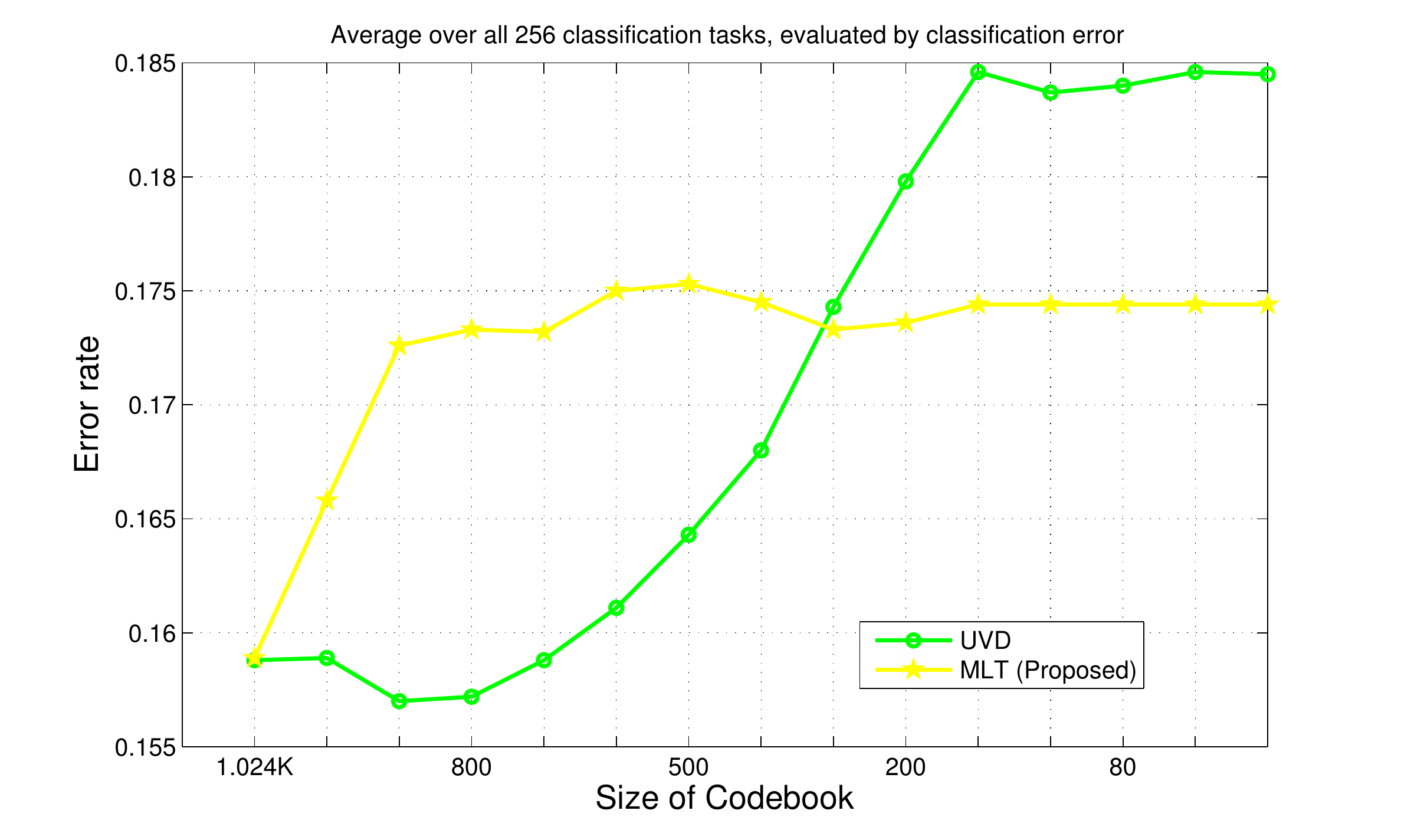}}
            \subfloat[]{ \includegraphics[height=43mm,width=50mm]{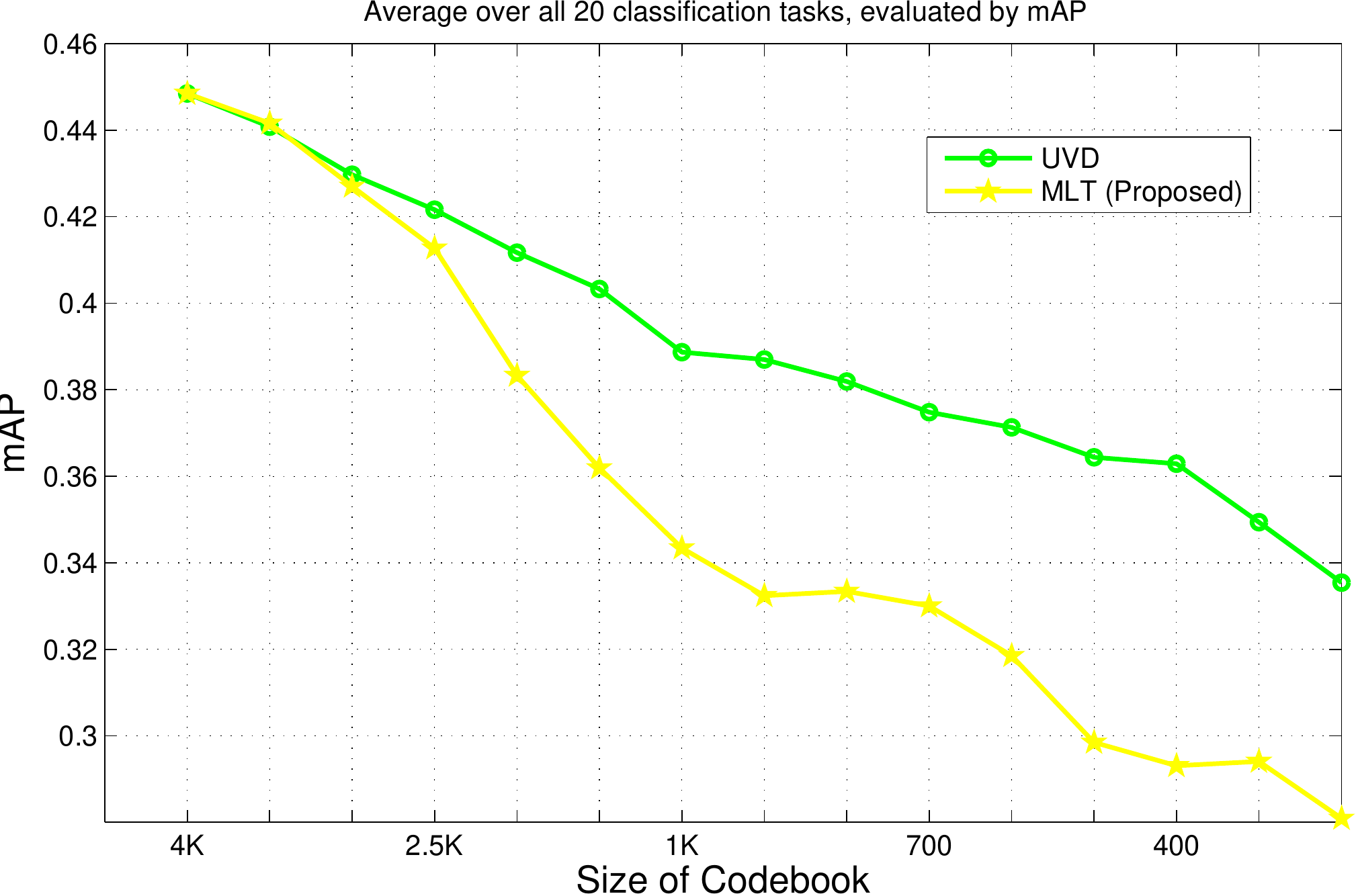}} 
            \subfloat[]{ \includegraphics[height=43mm,width=50mm]{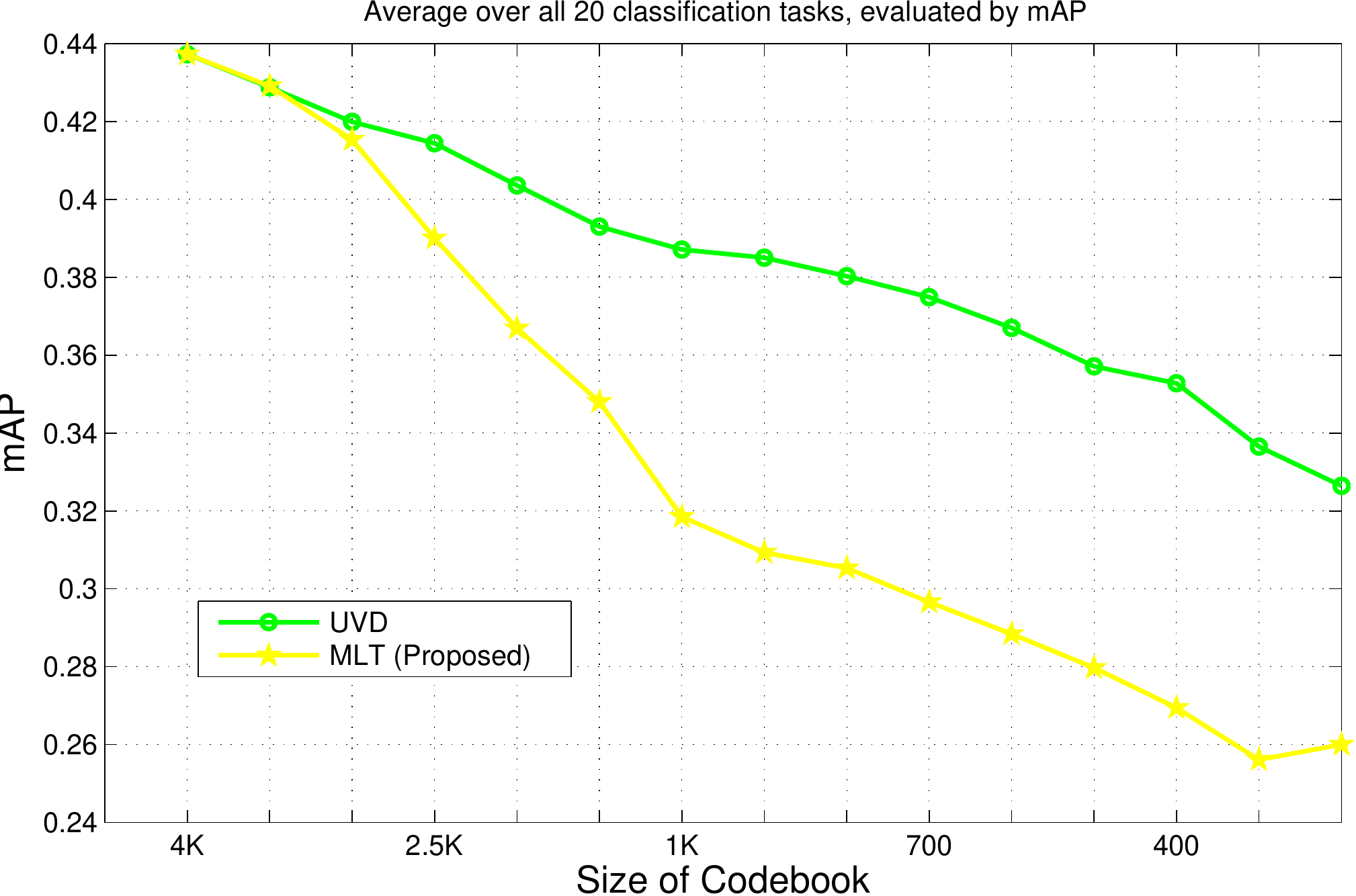}}

    \end{tabular}

    \caption{ Comparison of UVD and MLT on Caltech-256 (a), PASCAL 2007 (b) and PASCAL 2012 (c). 
    }
    \label{fig:MLT_Comparison_Basic}
    \end{figure*}

  \begin{figure*}[!ht]
  	\center
    \begin{tabular}{l}
            \subfloat[]{ \includegraphics[height=43mm,width=50mm]{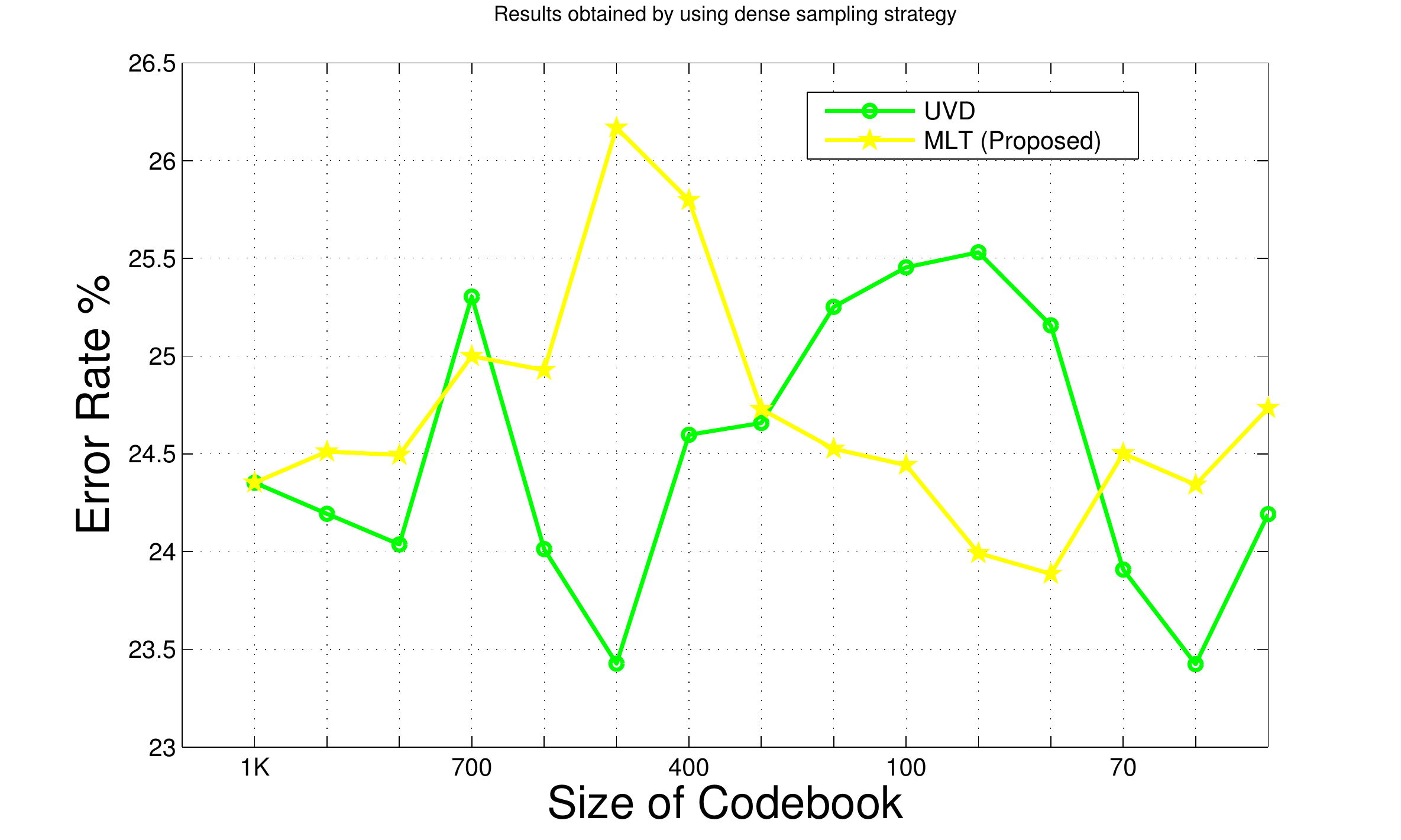}} 
            \subfloat[]{ \includegraphics[height=43mm,width=50mm]{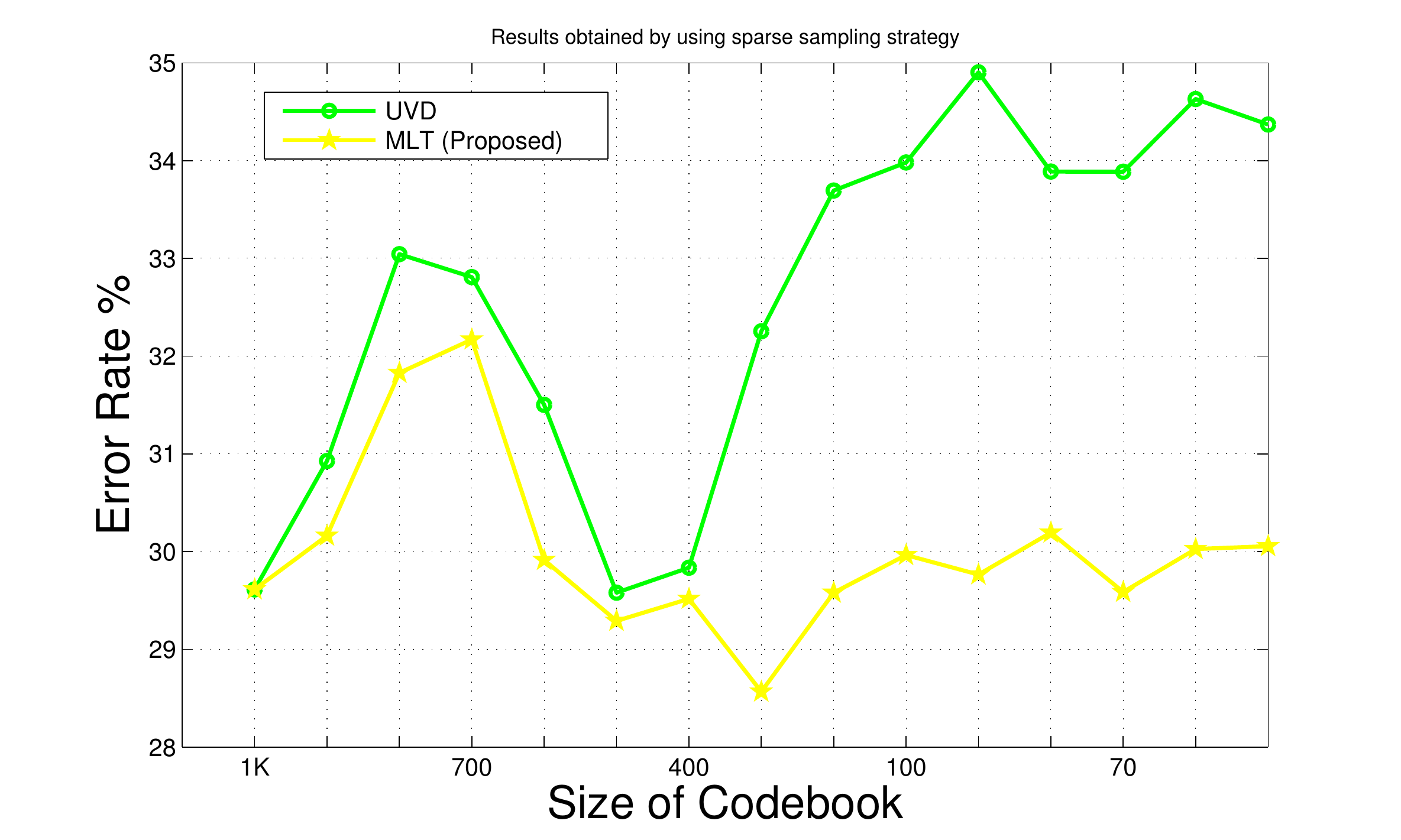}}
    \end{tabular}

    \caption{ Comparison of UVD and MLT on Graz02 with dense sampling and sparse sampling strategies. (a) The result obtained by using the dense sampling strategy. (b) The result obtained by using the sparse sampling strategy.
    }
    \label{fig:control_experiment}
    \end{figure*}

Compared with UVD, MLT only changes the distribution model from Gaussian distribution to Multinomial distribution. Thus the performance comparison between these two methods demonstrates the impact of using different distribution models. Recall that MLT is inspired from the fact that in text analysis multinomial distribution is more commonly adopted in the bag-of-words model. But is multinomial distribution still suitable for the visual words extracted from the bag-of-features model in visual recognition? To answer this question, in this section, we firstly compare UVD and MLT on four datasets. They are KTH, Caltech-256, PASCAL 2007 and PASCAL 2012. 

The performance comparison between UVD and MLT on KTH is shown in Figure \ref{fig:KTH_Comparison}. As seen from the average performance (Figure \ref{fig:KTH_Comparison} (a)) on six one-vs-rest tasks, MLT significantly outperforms UVD, especially when the codebook size is reduced to a small number. The same trend is observed on the two most difficult tasks: `running vs. the rest' in Figure \ref{fig:KTH_Comparison} (b) and `jogging vs. the rest' in Figure \ref{fig:KTH_Comparison} (c). For the result on Caltech-256 shown in Figure \ref{fig:MLT_Comparison_Basic}, we can see that UVD performs better at the beginning of the merging process, but when the codebook size is reduced to be less than 300, it is outperformed by MLT. For the result on Pascal 2007 and Pascal 2012, however, MLT performs much worse than UVD.

To explain the better performance of MLT over UVD on KTH dataset, we notice that among these four datasets, the feature extraction scheme used for KTH is different from that used for the other three datasets. In KTH, the local features are extracted from a set of detected interest points while in the other three datasets the local features are extracted in a dense spatial grid, namely, using a dense sampling strategy. Note that in the multinomial distribution model, the occurrence of words is assumed to follow the i.i.d property. However, for the dense sampling strategy, the neighboring sampling points are spatially close to each other. Due to the Markov property of images, the visual patterns within a neighborhood are often co-occurred. Thus, the neighboring local features and their quantized visual words can be highly correlated and the i.i.d assumption taken in the multinomial distribution model tends to be violated. In contrast, the strategy of extracting local features around interest points introduces much less correlation among visual words because interest points are usually spatially scattered. As a result, the multinomial distribution is more appropriate for modeling the histogram in KTH than that in the other three datasets. 

To further verify the above  interpretation, we apply both dense and sparse sampling strategies on Graz02 \footnote{For the simplicity of experiment, we employ the Graz02 dataset as the test benchmark because its size is relatively smaller than Caltech256, PASCAL 07 and PASCAL 12. Here, it is used to evaluate image-level classification performance here. } to create two datasets. Following the same experiment protocol used above, we obtain the performance comparison between UVD and MLT on both datasets, shown in Figure \ref{fig:control_experiment} (a)(b). It can be seen that the UVD and MLT show quite similar performance in the dataset obtained by using the dense sampling strategy while MLT significantly outperforms UVD in the dataset obtained by using the sparse sampling strategy. This is consistent with the observation made in KTH and supports the above interpretation.

From the above observation and discussion, the impact of the distribution model in our framework is clearly demonstrated: if the distribution model well represents the image representation, better performance can be obtained. In contrary, if the distribution model is inappropriate for modeling the image representation, the performance of the resultant merging algorithm will suffer. On KTH, the local feature extraction scheme makes the occurrence of visual words more independent of each other and in this case the visual words resemble the keywords in document analysis. Consequently, multinomial distribution becomes a better probabilistic model and MLT significantly outperforms UVD. In PASCAL VOC or Caltech-256, dense sampling strategy is adopted and multinomial distribution becomes inappropriate for modeling the resultant image representation. Consequently, MLT performs less satisfying in such cases. 

\subsection{The Evaluation of GMLE}
  \begin{figure*}[!ht]
    \begin{tabular}{l}
            \subfloat[]{ \includegraphics[height=43mm,width=50mm]{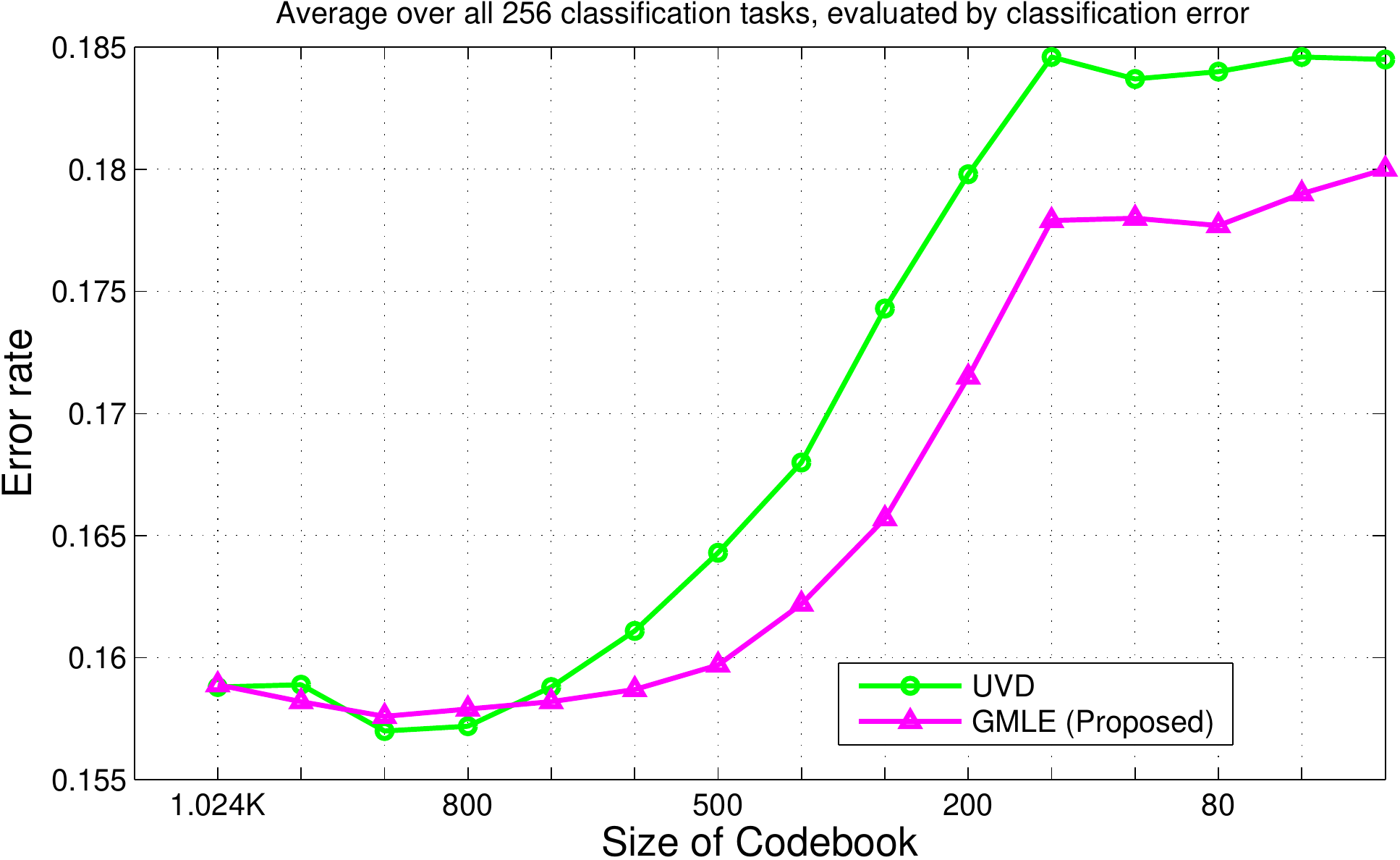}}
            \subfloat[]{ \includegraphics[height=43mm,width=50mm]{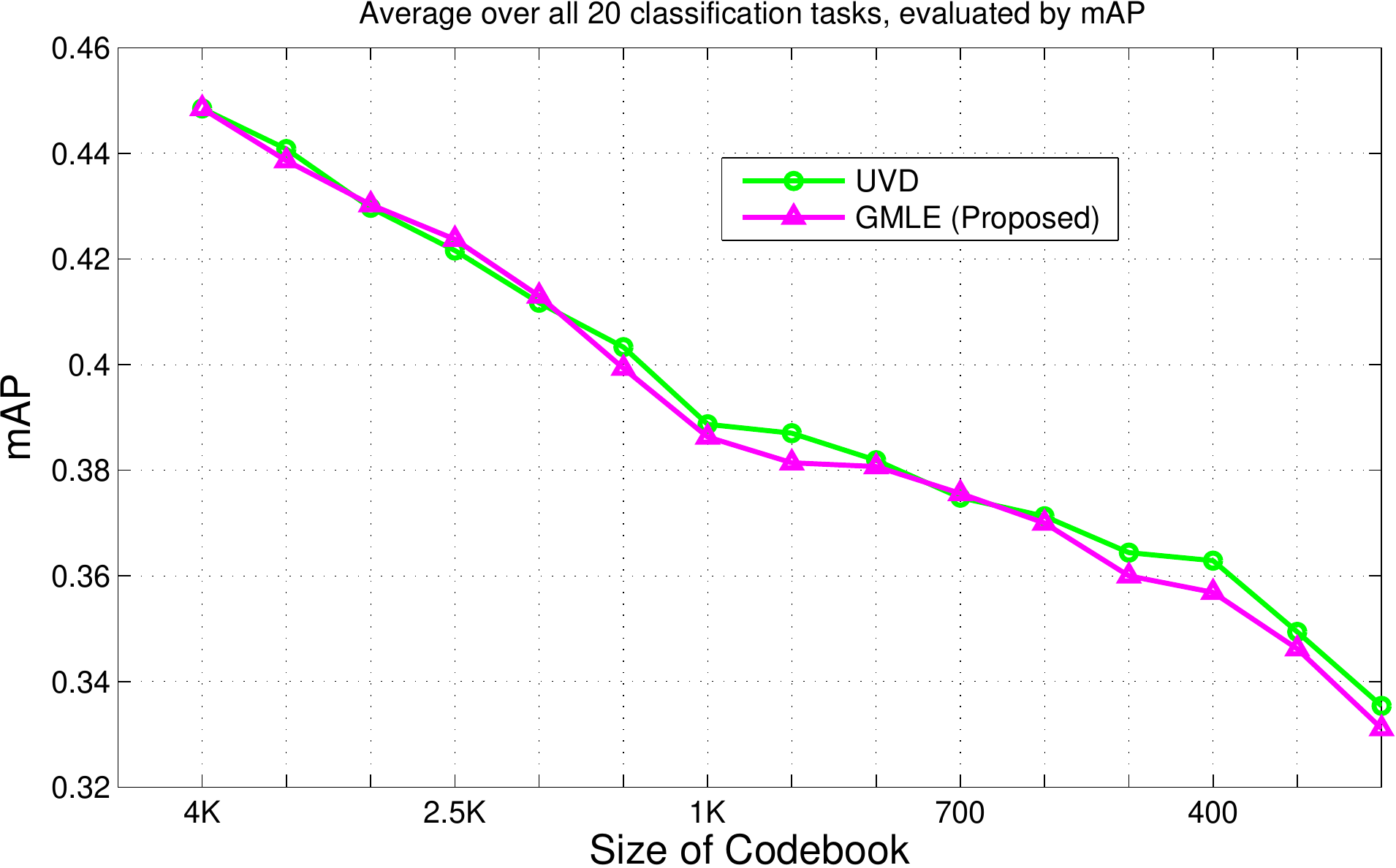}} 
            \subfloat[]{ \includegraphics[height=43mm,width=50mm]{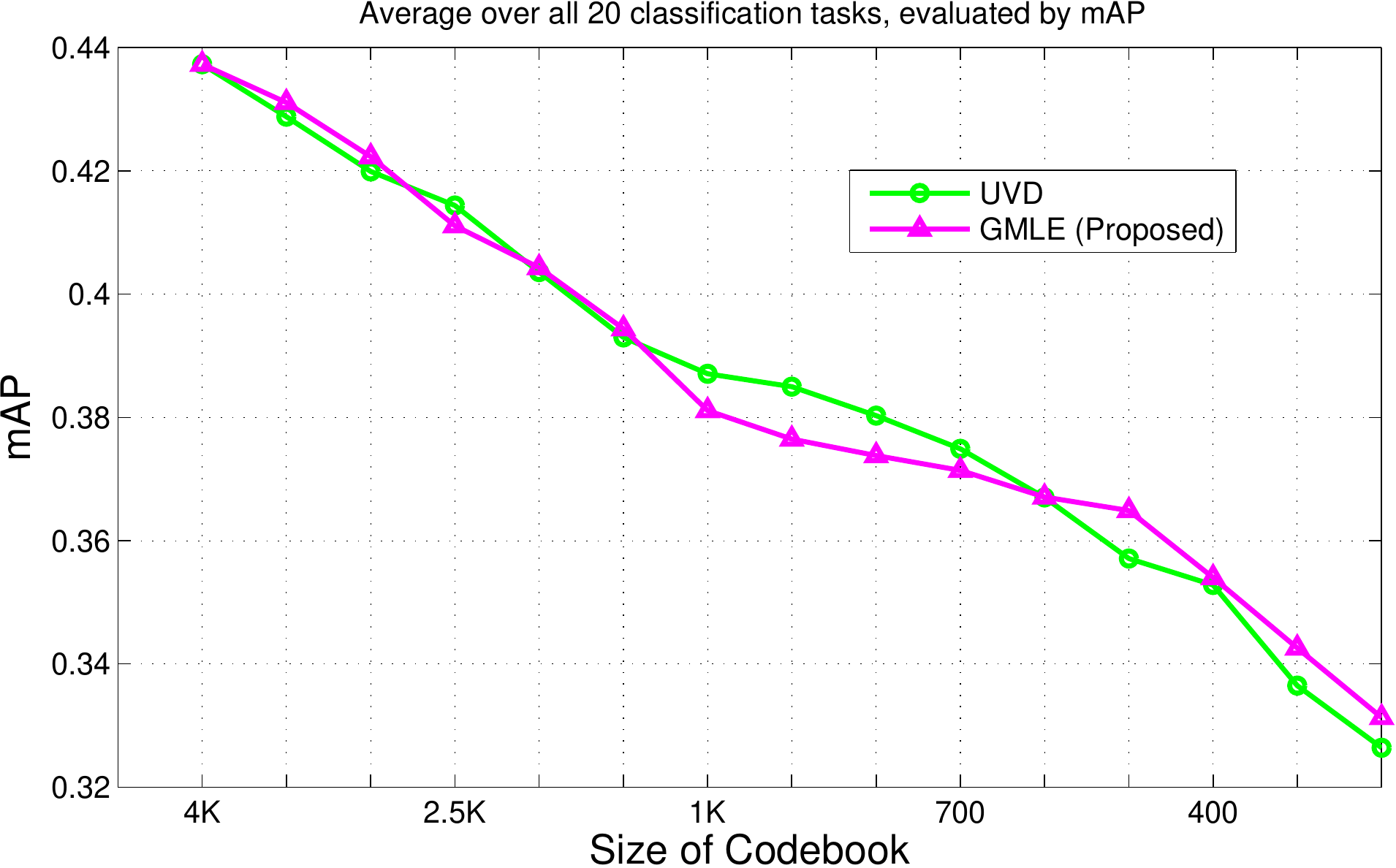}}

    \end{tabular}

    \caption{ Comparison of UVD and GMLE on Caltech-256, PASCAL 2007 and PASCAL 2012. 
    }
    \label{fig:GMLE_Comparison_Basic}
    \end{figure*}
Compared with UVD, GMLE only replaces the parameter estimation method with maximum likelihood estimate. Thus, from the comparison between GMLE and UVD, the importance of parameter estimation can be demonstrated. Figure \ref{fig:GMLE_Comparison_Basic} shows the comparison. As seen, in all three datasets (Caltech 256, PASCAL 2007 and PASCAL 2012) GMLE outperforms or at least performs equally well as UVD. This supports our claim that MLE can be comparable to or even better than the Bayesian method for our framework because it directly learns the model parameters from the training data rather than relying on an empirical choice of hyper-parameters as in UVD. 


\subsection{The Evaluation of MME}
%
  \begin{figure*}[!ht]
    \begin{tabular}{l}
            \subfloat[]{ \includegraphics[height=43mm,width=50mm]{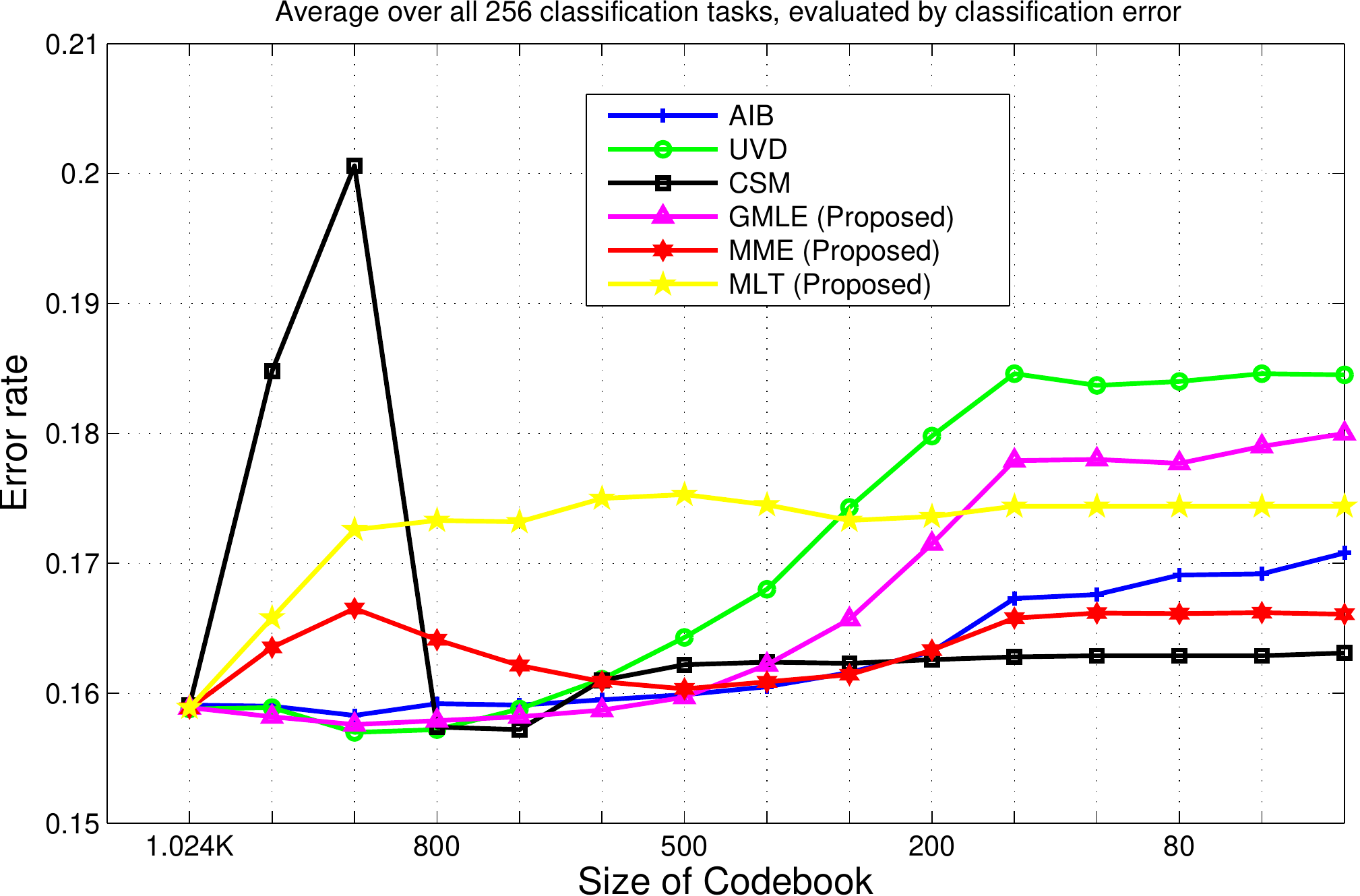}}
            \subfloat[]{ \includegraphics[height=43mm,width=50mm]{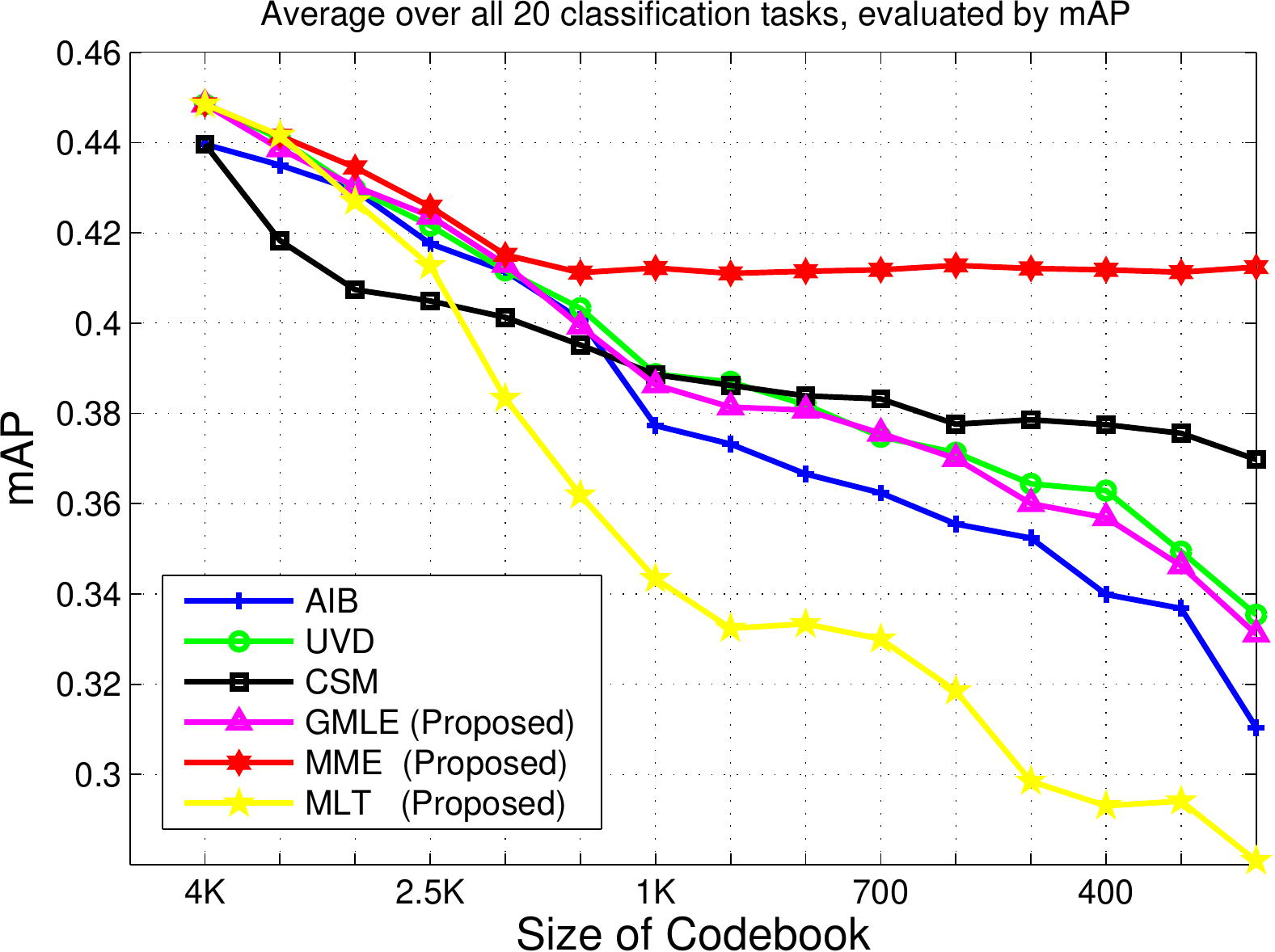}} 
            \subfloat[]{ \includegraphics[height=43mm,width=50mm]{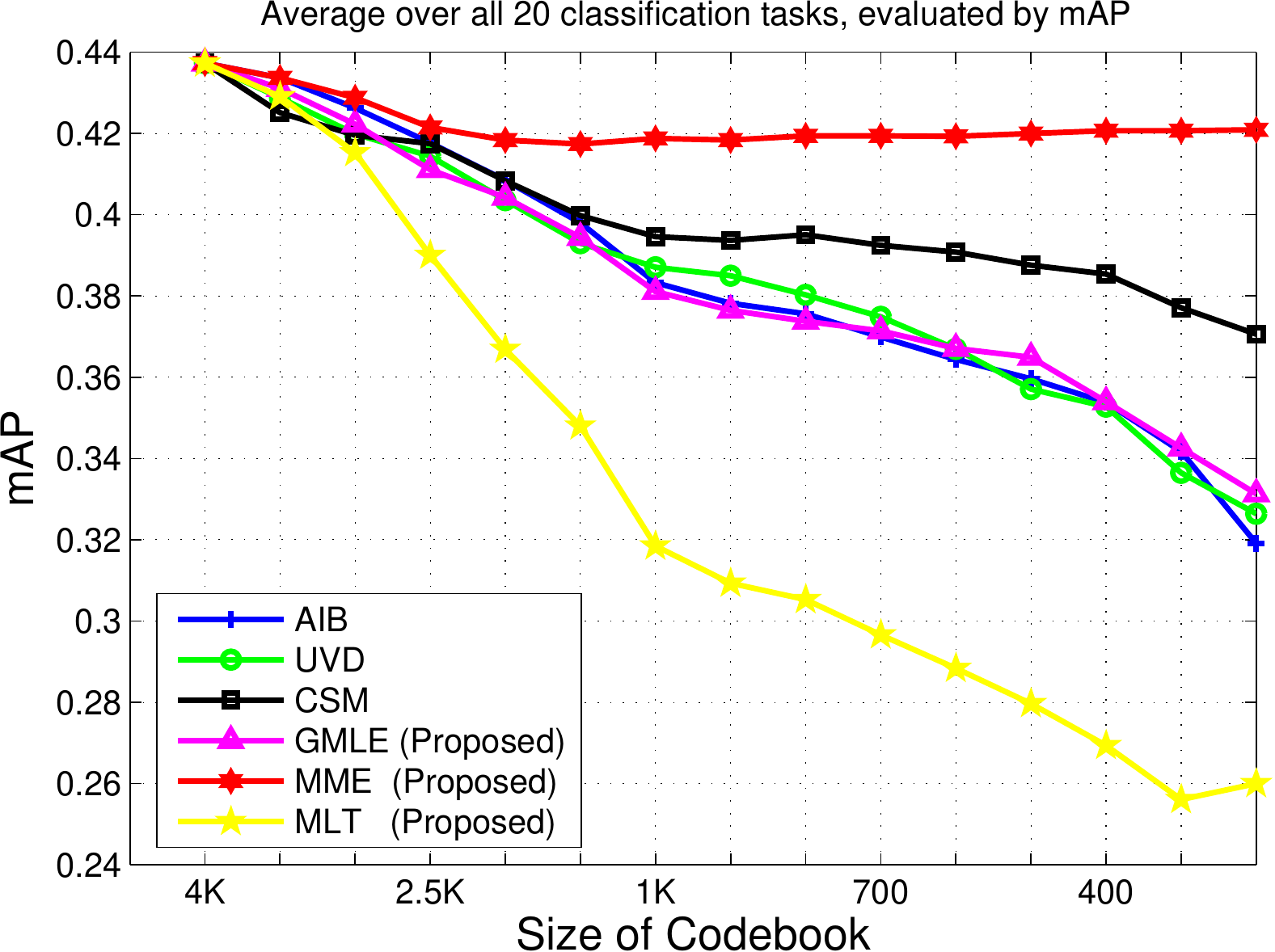}}

    \end{tabular}

    \caption{ Comparison of all six methods on Caltech-256, PASCAL 2007 and PASCAL 2012. 
    }
    \label{fig:All_Comparison_Basic}
    \end{figure*}

In this section, we compare the performance of MME against all the other five methods. MME adopts a more advanced parameter estimation method which incorporates the discriminative information into parameter estimation process. Thus it is expected to be better at maintaining the discriminative power of an initial codebook. In Figure \ref{fig:All_Comparison_Basic}, we evaluate the performance of MME on Caltech-256, PASCAL 2007 and PASCAL 2012. 

In Caltech-256, the performance of MME becomes the second best method when the codebook size is reduced to 200 \footnote{Generally speaking, for a supervised compact codebook creation method, the performance with a smaller codebook size is usually more important since the advantage of using compact codebook is more pronounced in such scenario.} and its difference from the best method is very marginal (less than 0.5\%). 

In the more challenging PASCAL 2007 and 2012 datasets, the advantage of MME is more clearly demonstrated. As seen in Figure \ref{fig:All_Comparison_Basic}, after a slight drop in the period when the codebook size is reduced from 4000 to 1500, its classification performance is steadily kept in the remaining course of merging process (from 1500 to 100). This is in a sharp contrast to the quick performance drop of the other merging methods. Compared with CSM -- the one achieving the second best performance -- the improvement can be as large as 4-5\%. This well demonstrates the advantage of using max-margin parameter estimation in our framework.   

Interestingly, MME adopts the multinomial distribution and it still achieves excellent performance on PASCAL datasets in which multinomial distribution may not be an accurate model. It seems that using supervised parameter estimation can compensate the disadvantage caused by choosing a less appropriate distribution.    

\subsection{Evaluation on the application of pixel-level detection}
In this section, we further compare the word merging algorithms on the pixel-wise object detection problem \cite{smartdictionary}. As indicated by \cite{smartdictionary}, to perform the pixel-wise detection, we need to calculate the histogram of visual words occurring within the region centered at each pixel and this can be time-consuming if we implement it directly. An efficient way is to leverage the integral histogram \cite{IntegralHistogram} to quickly compute the histogram for a given region. However, the memory usage and computational cost will increase linearly with the size of codebook. If we could reduce the codebook size without significantly sacrificing the classification performance, then a better trade-off between the performance and computational complexity could be achieved. Compact codebook created by word merging algorithms fits perfectly to this demand. 

In Table \ref{table:GrazDetection}, we compare the average detection performance obtained by applying different merging algorithms with respect to different codebook size. The performance is measured by EER (Equal Error Rate) as in \cite{smartdictionary}. As seen, MME achieves the overall best performance. It well maintains the discriminative power of the initial codebook. The EER achieved with a 10-word codebook is comparable to that obtained with the 1000-word initial codebook. Thus, by using this compressed codebook, we only need 1/100 computational cost and memory usage of that required in the direct implementation. 

  \begin{table*}[ht!]
        \begin{center}
        \caption{Comparison of EER for six merging methods on Graz02. }
        \label{table:GrazDetection}
            \begin{tabular}{llllllll}
            \hline\noalign{\smallskip}
 			codebook size &    1000 (initial) & 200   &    150   &   100   &   80   &   50   &    20  \\
            \noalign{\smallskip}
            \hline
            \noalign{\smallskip}
			 MME    & 0.621      & \bf 0.623    & \bf 0.620   & 0.614  & \bf 0.635 & \bf 0.621 & \bf 0.625  \\
			 AIB    & 0.621       & 0.604    & 0.598   & \bf 0.622  & 0.600 & 0.617 & 0.596  \\
			 CSM    & 0.621       & 0.588    & 0.587   & 0.609  & 0.612 & 0.599 & 0.587 \\
			 UVD    & 0.621       & 0.604    & 0.600   & 0.602  & 0.578 & 0.582 & 0.601 \\
			 MLT    & 0.621       & 0.573    & 0.571   & 0.584  & 0.581 & 0.605 & 0.596 \\
			 GMLE   & 0.621       & 0.601    & 0.569   & 0.585  & 0.607 & 0.609 & 0.572 \\
            \hline
            \end{tabular}
        \end{center}
     \end{table*}

\section{Discussion on the impact of scaling factor}
   Throughout our experiments, we set the scaling factor $\eta$ to a small value (0.01) in our MME method. As discussed in the Appendix A, a small $\eta$ ensures that the estimated probability values lie between 0 and 1. However, among those possible values of $\eta$ that guarantee valid probability estimates, we still have many choices. Then a question arises, what is the impact of the value $\eta$ on the performance of MME? In this section, we discuss this issue with both empirical evaluation and theoretical analysis.

\begin{figure}
    \centering
            \includegraphics[height=35mm]{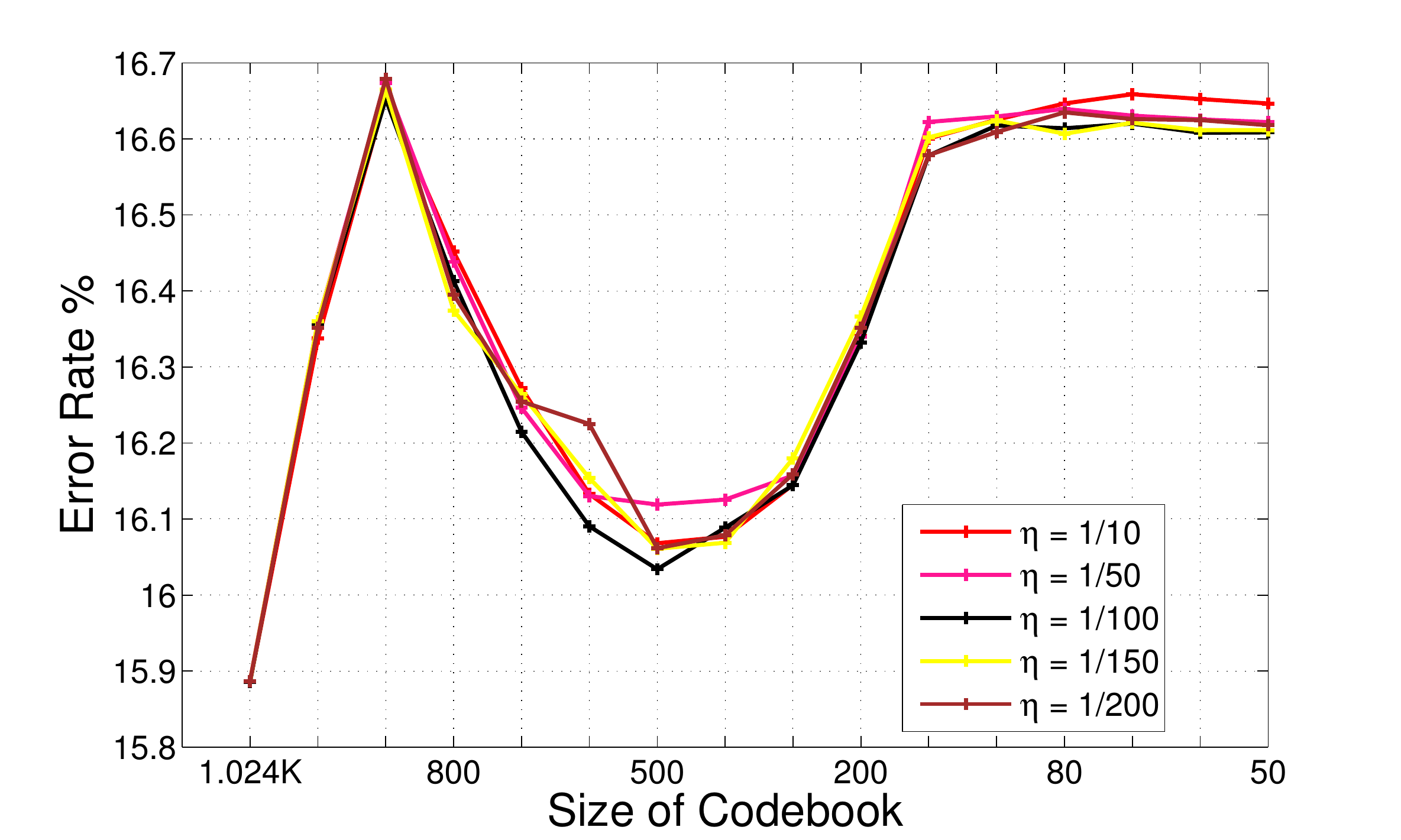}
    \caption{
    The impact of the scaling factor on Caltech-256. 
    }
    \label{fig:Scaling Factor}
\end{figure}

	For the empirical evaluation, we re-evaluate the performance of MME on Caltech256 with different scaling factors. We test a range of scaling factors -- $\{1/10,1/50,1/100,1/150,1/200 \}$ and show the result in Fig. \ref{fig:Scaling Factor}. From the result, it is clear that the performance obtained by using different scaling factors is very similar. This suggests that the choice of scaling factor has little impact on the performance of MME once it is set to a relatively small value.
	
	To further justify our empirical observation, we analyse this issue from the theoretical aspect. As discussed in Section \ref{sect:MME_solution}, once the max-margin parameter estimation is completed at each level, the identification of the word pair follows the same criterion as in AIB, that is, the best word pair should maximize the merging criterion:
\begin{align}
	(r^*,s^*) & = \mathop{\mathrm{argmax}}_{r,s} \mathcal{J}_{AIB}(\mathcal{H}_{r,s}^{t}) \nonumber \\
		      & = \mathop{\mathrm{argmin}}_{r,s} \mathcal{J}_{AIB}(\mathcal{H}^{t-1}) - \mathcal{J}_{AIB}(\mathcal{H}_{r,s}^{t}) \nonumber \\
		      & = \mathop{\mathrm{argmin}}_{r,s} cost(r,s),
\end{align}
where $\mathcal{H}_{r,s}^{t}$ denotes the training histograms obtained after merging the $r$th and $s$th words. We can show that (see Appendix B) when the scaling factor $\eta$ is small, $cost(r,s)$ can be approximated by
\begin{align}\label{approximation}
	cost(r,s) \approx \eta t_{r,s},
\end{align}
where $t_{r,s}$ is a term which does not involve $\eta$. In other words, Eq. (\ref{approximation}) suggests that the scaling factor just scales the cost term and the relative relationship between the costs of different pairs is almost unaffected. Thus, the identified merging pair $(r^*,s^*)$ tends to remain the same even though different scaling factors are used.

\section{Conclusion}
	This paper presents a generalized probabilistic framework to both unify existing visual words merging criteria and induce new criteria for compact codebook construction. The key insight of this framework is that different merging criteria can be realized by changing two key factors identified in the proposed framework, that is, the function used to model the class-conditional distribution and the method to handle parameter estimation for the distribution model. By appropriately setting these two factors, we not only recover the existing merging criteria but also create three new criteria, named as MLT, GMLE and MME. Through the experimental comparison between these three criteria and the existing ones, we made three main discoveries: 1) The appropriateness of the distribution model choice could have significant impact on the performance of a word merging criterion. 2) Besides the Bayesian method, MLE and MME are also good parameter estimation methods in our framework. MLE is comparable or even better than the Bayesian method since it does not need to empirically set the hyper-parameter. 3) MME achieves the overall best performance, demonstrating the power of using the max-margin objective to perform parameter estimation. In our future work, we will further study this framework for more visual learning tasks, for example, instead of focusing on classification, we could extend the proposed framework for creating compact codebook in metric learning setting. Also, the computational efficiency of the proposed MME will be addressed to handle higher dimensional image representation.

\section{Appendix A: Discussion on the two stage solution for Eq.(\ref{final_problem})}
The two-stage optimization method shown in Eq.(\ref{final_problem}) is valid if for a solution $\mathbf{w}^*$ and $b^*$ obtained in the first stage, we can find a scaling factor $\eta$ which makes the last three constraints in Eq.(27) satisfied. This is because the solution $\mathbf{w}^*$ and $b^*$ attained without the last three constraints always gives a lower or equal objective value than the one which considers these additional constraints. Thus, $\mathbf{w}^*$ and $b^*$ will be the optimal solution if the last three constraints can be automatically satisfied by tuning the scaling factor. To examine when this is true, we first derive the solution for $P(v_j|c)$ and $P(c)~~c = 1,-1$ according to constraints:
\begin{align}
	& \eta w_j = \mathrm{log}(\frac{P(v_j|c = 1)}{P(v_j|c=-1)}) \nonumber \\
	& \eta b = \mathrm{log}(\frac{P(c= 1)}{P(c=-1)}) \nonumber \\
    & \sum_{c = -1,+1} P(v_j|c)P(c) = P(v_j), ~~~ \sum_{c = -1,+1} P(c) = 1 \nonumber.
\end{align}
The solutions of above equalities can be worked out as: 
\begin{align}\label{prob_solution}
	& P(c = 1) = \frac{\exp(\eta b )}{1+\exp(\eta b)} \nonumber \\
    & P(c = -1) = \frac{1}{1+\exp(\eta b)} \nonumber \\
	& P(v_j|c = -1) = P(v_j) \frac{1+\exp(\eta b)}{1+\exp(\eta b)\exp(\eta w_j)} \nonumber \\
    & P(v_j|c = 1) = P(v_j) \frac{(1+\exp(\eta b))\exp(\eta w_j)}{1+\exp(\eta b)\exp(\eta w_j)}. \nonumber \\
\end{align}
From the above solutions, we could see that $P(c=1)$ and $P(c=-1)$ are always between 0 and 1. However, the solution of  $P(v_j|c = -1)$ and $ P(v_j|c = 1)$ could be greater than 1 because the term $\frac{1+\exp(\eta b)}{1+\exp(\eta b)\exp(\eta w_j)}$ and $\frac{(1+\exp(\eta b))\exp(\eta w_j)}{1+\exp(\eta b)\exp(\eta w_j)}$ can be larger than 1. However, we noticed that when $\eta \rightarrow 0$, these two terms will approach 1. Meanwhile, since $P(v_j)$  is calculated via the MLE method, that is:
\begin{align}
	P(v_j) = \frac{\sum_{i|\mathbf{h}_i \in \mathcal{D}} h_{ij}}{\sum_{j=1}^{t}\sum_{i|\mathbf{h}_i \in \mathcal{D}} h_{ij}} ,
\end{align}
where $\mathcal{D}$ denotes the whole training set. Recall that $t$ is the compact codebook size and generally $t$ is much larger than 2. This will make the numerator much smaller than the denominator. Thus,  $P(v_j)$ is usually much smaller than 1. Hence, in practice, $P(v_j)$ will greatly scale down $\frac{1+\exp(\eta b)}{1+\exp(\eta b)\exp(\eta w_j)}$ and $\frac{(1+\exp(\eta b))\exp(\eta w_j)}{1+\exp(\eta b)\exp(\eta w_j)}$ terms and make $P(v_j|c)$ less than 1. This justifies our solution for the problem in Eq. (\ref{final_problem}). 

\section{Appendix B: The derivation of Eq.(\ref{approximation}) }

It is straightforward to derive $cost(r,s)$ as:
\begin{align}
	&cost(r,s) = \nonumber \\
	& \sum_{c=1}^{C}\sum_{j\in \{r,s\}} P(v_j,c)\left( \log \frac{P(v_j|c)}{P(v_j)} - \log \frac{P(v_r|c)+P(v_s|c)}{P(v_r)+P(v_s)} \right).
\end{align}  
According to Eq. (\ref{prob_solution}), $P(c)$ and $P(v_j|c)$ are functions of $w_j \eta$ and $b \eta$. In other words, $cost(r,s)$ can be seen as a function of the vector input $ \mathbf{x} = (w_r \eta, w_s \eta, b \eta)^T$. When $\eta$ is small, $cost(r,s)$ ($cost(\mathbf{x})$) can be approximated by its first order Taylor expansion at point 0, that is,
\begin{align}
	cost(\mathbf{x}) & \approx cost(\mathbf{0}) + \langle \frac{\partial cost(\mathbf{0})}{\partial{\mathbf{x}}},\mathbf{x}\rangle  \nonumber \\
		& = cost(\mathbf{0}) + \eta \langle\frac{\partial cost(\mathbf{0})}{\partial{\mathbf{x}}},(w_r , w_s , b )\rangle  \nonumber \\
		& =  cost(\mathbf{0}) + \eta t_{r,s},
\end{align} 
where $\langle\cdot,\cdot\rangle$ denotes the inner product and $t_{r,s}$ is a term that does not involve $\eta$. From Eq. (\ref{prob_solution}), it can be seen that if $(w_r \eta, w_s \eta, b \eta) = \mathbf{0}$, $P(v_j|c) = P(v_j)$ and $P(c) = \frac{1}{2} ~~ c = 1 ,-1$. Then, it is easy to verify that $cost(\mathbf{0}) = 0$. Thus, we could arrive at Eq. (\ref{approximation}).

\IEEEpeerreviewmaketitle

\bibliographystyle{ieee}
\bibliography{CVPR2011}

\end{document}